\documentclass[conference]{IEEEtran}
\IEEEoverridecommandlockouts

\ifCLASSINFOpdf
\else
\fi
\usepackage{hyperref}
\usepackage{url}
\usepackage{amsmath}
\usepackage{amsthm}
\usepackage{caption}
\usepackage{amssymb}
\usepackage{algorithm}
\usepackage{algpseudocode}
\usepackage{wrapfig}
\usepackage{varwidth}
\usepackage{lipsum}
\usepackage{makecell}
\usepackage{listings}
\usepackage{xcolor}
\usepackage{multirow}
\usepackage{graphicx}
\usepackage{subcaption}
\usepackage{hyperref}
\usepackage{colortbl}

\definecolor{red}{RGB}{178,33,33}
\definecolor{blue}{RGB}{64,105,224}

\begin{document}
%
\title{Causal Structure Learning \\ Supervised  by Large Language Model}

\author{
\IEEEauthorblockN{Taiyu Ban\qquad Lyuzhou Chen\qquad Derui Lyu\qquad Xiangyu Wang$^*$\qquad Huanhuan Chen$^*$}
\IEEEauthorblockA{
\textit{School of Computer Science and Technology, University of Science and Technology of China}\\
\texttt{\{\href{mailto:banty@mail.ustc.edu.cn}{banty},\href{mailto:clz31415@mail.ustc.edu.cn}{clz31415},\href{mailto:drlv@mail.ustc.edu.cn}{drlv}\}@mail.ustc.edu.cn
\{\href{mailto:sa312@ustc.edu.cn}{sa312},\href{mailto:hchen@ustc.edu.cn}{hchen}\}@ustc.edu.cn}}
\thanks{$^*$ These authors are corresponding authors.}}


%


\maketitle


\begin{abstract}
Causal discovery from observational data is pivotal for deciphering complex relationships. Causal Structure Learning (CSL), which focuses on deriving causal Directed Acyclic Graphs (DAGs) from data, faces challenges due to vast DAG spaces and data sparsity. 
The integration of Large Language Models (LLMs), recognized for their causal reasoning capabilities, offers a promising direction to enhance CSL by infusing it with knowledge-based causal inferences. 
However, existing approaches utilizing LLMs for CSL have encountered issues, including unreliable constraints from imperfect LLM inferences and the computational intensity of full pairwise variable analyses.
In response, we introduce the Iterative LLM Supervised CSL (ILS-CSL) framework. 
ILS-CSL innovatively integrates LLM-based causal inference with CSL in an iterative process, refining the causal DAG using feedback from LLMs. 
This method not only utilizes LLM resources more efficiently but also generates more robust and high-quality structural constraints compared to previous methodologies.
Our comprehensive evaluation across eight real-world datasets demonstrates ILS-CSL's superior performance, setting a new standard in CSL efficacy and showcasing its potential to significantly advance the field of causal discovery. The codes are available at \url{https://github.com/tyMadara/ILS-CSL}.
\end{abstract}


%
\IEEEpeerreviewmaketitle


\section{Introduction}
\label{sec_intro}

Causal discovery from the observed data is pivotal in understanding intricate relationships across various domains. 
Central to this endeavor is Causal Structure Learning (CSL), aiming to construct a causal Directed Acyclic Graph (DAG)\footnote{In a causal DAG, each edge represents a direct causal link between its nodes.} from observed data \cite{pearl2009causality}. We adopt causal Bayesian Networks (BNs) as the causal graphical model, renowned for effectively modeling intricate real-world variable relationships \cite{ellis2008learning}.

The recovery of high-quality causal BNs faces significant challenges. Firstly, there is the issue of the super-exponential increase in the DAG space as the number of variables grows \cite{chickering1996learning,kitson2023survey}. Additionally, real-world data is typically sparse and insufficient for accurately representing the true probability distributions \cite{morgan2015counterfactuals}. Furthermore, the orientation of edges in a BN cannot be fully deduced from the observed data alone due to the presence of equivalent DAGs \cite{chickering2002optimal}. In summary, CSL, when reliant solely on observed data, encounters both practical and theoretical limitations.

Given these inherent limitations, the integration of prior knowledge to constrain specific structures becomes important for reliable causal discovery~\cite{chen2016learning,amirkhani2016exploiting}. While promising, this approach has been limited by the high costs and time associated with expert input~\cite{constantinou2023impact}. However, the advent of Large Language Models (LLMs) has ushered in a new frontier. Recent studies have underscored the capabilities of LLMs in causal reasoning, positioning them as a valuable and readily accessible resource for knowledge-based causal inference~\cite{kiciman2023causal,nori2023capabilities,chen2023mitigating}.

K{\i}c{\i}man $et$ $al$. have shown that Large Language Models (LLMs) are effective in determining causality direction between pairs of variables, outperforming even human analysis in this respect \cite{kiciman2023causal}. However, other studies highlight LLMs' limitations in constructing causal DAGs from sets of variables, not satisfying even in small-scale contexts \cite{tu2023causal,long2023can}. This difficulty mainly stems from the inherent complexity in inferring detailed causal mechanisms, such as establishing the relative directness of causes for an effect, a task that often exceeds simple knowledge-based inference.

In response to these challenges, recent studies have begun integrating LLM-derived causal knowledge with data analysis to enhance causal discovery. For example, Ban $et$ $al$. \cite{ban2023query} utilize LLMs to discern the presence of causal links among variables, subsequently applying ancestral constraints to structure learning \cite{chen2016learning}. This approach yields improvements in learning causal structures from data for smaller-scale problems, but it encounters difficulties with larger datasets due to inaccuracies in the LLM-derived constraints, as evidenced in Table \ref{tab:ban_cmp}. As an alternative, Vashishtha $et$ $al$. \cite{vashishtha2023causal} employ a detailed, pair-based prompting strategy with a voting system to determine reliable prior knowledge. Regretably, the authors fail to show the effectiveness on the larger-scale datasets, likely limited by the complexity and computational demands of the prompt process, which requires \(\binom{N}{2}\) LLM inferences with $N$ denoting the variable count.


\begin{table*}[!h]
\setlength{\tabcolsep}{5pt}
\small
\caption{SHD$\downarrow$ and constraint quality of the ancestral constraint-based CSL driven by GPT-4, reported in the work \cite{ban2023query}.}
\label{tab:ban_cmp}
\begin{center}
\begin{tabular}{c|cc|cc|cc|cc|cc|cc|cc|cc}
\hline
Dataset& \multicolumn{2}{c|}{\begin{tabular}[c]{@{}c@{}}Cancer\\ 5 nodes\end{tabular}} & \multicolumn{2}{c|}{\begin{tabular}[c]{@{}c@{}}Asia\\ 8 nodes\end{tabular}} & \multicolumn{2}{c|}{\begin{tabular}[c]{@{}c@{}}Child\\ 20 nodes\end{tabular}} & \multicolumn{2}{c|}{\begin{tabular}[c]{@{}c@{}}Insurance\\ 27 nodes\end{tabular}} & \multicolumn{2}{c|}{\begin{tabular}[c]{@{}c@{}}Alarm\\ 37 nodes\end{tabular}} & \multicolumn{2}{c|}{\begin{tabular}[c]{@{}c@{}}Mildew\\ 35 nodes\end{tabular}} & \multicolumn{2}{c|}{\begin{tabular}[c]{@{}c@{}}Water\\ 32 nodes\end{tabular}} & \multicolumn{2}{c}{\begin{tabular}[c]{@{}c@{}}Barley\\ 48 nodes\end{tabular}} \\ \hline
Data size& 250& 1000& 250& 1000 & 500 & 2000 & 500 & 2000 & 1000& 4000 & 8000& 32000 & 1000& 4000& 2000& 8000\\ \hline
MINOBSx& 3.0& 1.8& 4.2& 2.5& \cellcolor{gray!25}9.5& \cellcolor{gray!25}5.3 & 25.7& \cellcolor{gray!25}\textbf{15.0} & \cellcolor{gray!25}9.5& \cellcolor{gray!25}6.5 & \cellcolor{gray!25}\textbf{22.8}& \cellcolor{gray!25}\textbf{21.0} & \cellcolor{gray!25}\textbf{62.3} & \cellcolor{gray!25}\textbf{53.7} & \cellcolor{gray!25}\textbf{47.0} & \cellcolor{gray!25}\textbf{33.7} \\
+GPT-4 & \textbf{0.5}& \textbf{0.0}& 2.2& 0.3& \cellcolor{gray!25}10.5& \cellcolor{gray!25}7.8 & \textbf{24.5} & \cellcolor{gray!25}16.2& \cellcolor{gray!25}12.3& \cellcolor{gray!25}8.8 & \cellcolor{gray!25}40.5& \cellcolor{gray!25}21.5 & \cellcolor{gray!25}66.7 & \cellcolor{gray!25}55.7 & \cellcolor{gray!25}52.0 & \cellcolor{gray!25}54.5 \\ \hline
CaMML & 2.0& 2.5& 3.5& 2.2& 6.0 & 1.0& 34.3& 31.7 & 11.0& 8.2& \cellcolor{gray!25}48.2& 62.2& 59.0& 53.2& \cellcolor{gray!25}81.5 & \cellcolor{gray!25}81.2 \\
+GPT-4 & 2.0& 1.3& \textbf{0.2} & \textbf{0.0} & \textbf{4.7} & \textbf{1.0} & 27.0& 22.2 & \textbf{6.0} & \textbf{3.0} & \cellcolor{gray!25}49.2& 60.0& 58.7& 48.3& \cellcolor{gray!25}82.2 & \cellcolor{gray!25}82.3
\\\hline
\makecell[c]{T / F} &\multicolumn{2}{c|}{5 / 0}  &\multicolumn{2}{c|}{9 / 0} &\multicolumn{2}{c|}{8 / 2} &\multicolumn{2}{c|}{10 / 0} &\multicolumn{2}{c|}{20 / 1} &\multicolumn{2}{c|}{9 / 6} &\multicolumn{2}{c|}{5 / 3} &\multicolumn{2}{c}{17 / 7} \\ \hline
\end{tabular}
\end{center}
\begin{flushleft}
The bold SHD is the best performance in each dataset.
The cell highlighted in gray indicates a degraded performance by integrating LLM-derived causal knowledge.
The row `T / F' represents the  number of correct LLM-derived structrual constraints (T) and that of erroneous ones (F).
\end{flushleft}
\end{table*}

In response to the challenges, we introduce a simple but effective strategy, named \underline{i}terative \underline{L}LM \underline{s}upervised CSL framework (ILS-CSL). 
Contrasting with prior methodologies that deploy LLMs and CSL separately, ILS-CSL uniquely focuses LLMs on verifying direct causal relationships already suggested by the data.
Specifically, ILS-CSL employs LLMs to validate the accuracy of edges in the learned causal DAG, with an iterative process fine-tuning CSL based on LLM feedback. 
The iteration concludes when the LLM-based inferences and data-driven CSL align within the established causal structure.
This innovative integration of LLMs into the CSL process offers significant enhancements to the task, as outlined below.

\begin{enumerate}
\item \textbf{Powerful Structural Constraints:} ILS-CSL transforms the causal inferences made by LLMs into structural constraints explicitly indicating the \textit{edge existence or absence}.
The edge-level constraint is more powerful than its path-level counterpart (ancestral constraint) in improving CSL\footnote{When an ancestral constraint is correctly identified, CSL might still recover a path that includes erroneous edges. In contrast, specifying the existence of an edge directly ensures accuracy, as it cannot be misinterpreted.}, with less risk\footnote{An incorrect ancestral constraint inevitably introduces at least one erroneous edge.}.
Please see Section \ref{sec_prelimianry_prior} for further discussions.
\item \textbf{Mitigation of Prior Errors:} ILS-CSL markedly diminishes the count of erroneous constraints, all while harnessing identical LLM resources. The reduction is theoretically by a factor of $O(N)$, estimated as $1.8(N-1)$, compared to the full inference on pairwise variables. Please refer to Section \ref{sec_theorem} for detailed estimation.
\item \textbf{Efficient Causal Inference with LLM:} 
ILS-CSL decreases the number of pairwise variable inferences from \( \binom{N}{2} \) to about \( O(N) \), as the LLM inference is restricted in the edges of of causal DAG\footnote{Given that the causal DAG is usually sparse, the number of edges $|E|$ is typically estimated as $O(N)$.}.
Such reduction makes the process more manageable and enhances the scalability of the framework.
\end{enumerate}

ILS-CSL has shown consistent improvement in data-driven CSL across all scales of the dataset used in the previous study \cite{ban2023query}. It effectively leverages various backbone causal discovery algorithms and demonstrates superior performance, especially as the number of variables increases. These results underscore ILS-CSL's significant potential for facilitating complex causal discovery tasks in real-world scenarios.




\section{Related Work}
This section discusses the emerging interest in the use of Large Language Models' (LLMs) common sense for understanding causal knowledge. It particularly focuses on the ways this knowledge is being harnessed in causal discovery.

\subsection{LLM-based Causal Discovery}

Recent advancements in LLM-based causal discovery primarily focus on assessing the inherent capabilities of LLMs \cite{willig2022can,liu2023evaluating}. Long $et$ $al$. \cite{long2023can} have tested LLMs' ability to generate simple causal structures, typically with sets of 3-4 variables. In a specialized domain, a study \cite{tu2023causal} investigates LLMs' effectiveness in discerning causal relationships within medical pain diagnosis, though the findings were somewhat inconclusive.

Kıcıman $et$ $al$. \cite{kiciman2023causal} have made strides in optimizing LLM performance for causal analysis by developing more refined prompting techniques. Their work assesses LLMs across a range of causal tasks, revealing notable performance in pairwise causal discovery \cite{hoyer2008nonlinear} and counterfactual inference \cite{frohberg2021crass}, even outperforming human analysis in certain aspects. Additionally, they have enhanced LLMs' capacity to identify causal structures in datasets concerning medical pain diagnosis. However, despite these advancements, a significant gap persists between the quality of causal DAGs generated by LLMs and those derived from data-based algorithms. These findings highlight the potential of LLM-based causal knowledge, yet they also underscore the importance of integrating data in uncovering genuine causal mechanisms.

\subsection{Integration of LLM in Data-based Causal Discovery}
A recent work first introduces LLM in causal discovery from data~\cite{ban2023query}.
Recognizing LLMs' limitations in differentiating indirect from direct causality, they applied ancestral constraints based on LLM-generated statements about the existence of causal relationships between variable pairs. 
The authors prompted the LLM with a complete set of variables, seeking the most confident causal assertions. 
However, when presented with numerous variables, LLM struggles to provide results that align with causal structures. 
This complexity leads to a decrease in the accuracy of causal statements as the number of variables increases, as demonstrated in Table \ref{tab:ban_cmp}.
Moreover, we observe that the LLM also fails to make comprehensive causal analyses in larger scale datasets as would be possible with individual prompts for each pair of variables. 

Motivated by this work,  Vashishtha $et$ $al$. \cite{vashishtha2023causal} adopted a more targeted method. They individually prompted the LLM for causal relationships between each variable pair and implemented a voting strategy to deduce ordering constraints. 
These constraints, although weaker than ancestral constraints (see Section \ref{sec_prelimianry_prior} for illustrations), offer more precise structural guidance for causal discovery. 
Their methodology demonstrates notable improvements across seven real-world datasets. However, the largest dataset examined containes only 23 nodes, leaving the approach's effectiveness in more complex scenarios untested.


\section{Preliminaries}

We begin by introducing the task of causal structure learning (CSL) on causal Bayesian Networks (BNs) and subsequently discuss the integration of structural constraints.

\subsection{Causal Bayesian Network}
A Bayesian Network (BN) is a probabilistic graphical model that uses a Directed Acyclic Graph (DAG) to represent conditional dependencies among a set of variables, thus defining their joint probability distribution. For a set of variables \( X = \{X_1, X_2, ..., X_n\} \) in a BN \( \mathcal{G} \), the joint probability distribution is given by:
\[
P(X_1, X_2, ..., X_n) = \prod_{i=1}^{n} P(X_i \mid \textbf{Pa}_i^{\mathcal{G}})
\]
\( \textbf{Pa}_i^{\mathcal{G}} \) denotes the parent nodes of \( X_i \) in the DAG. It's important to note that an edge in a BN does not inherently imply a causal relationship \cite{pearl2009causality}. 
A BN representing a joint probability distribution can be constructed using any variable ordering. However, the causal order of variables, indicating cause and effect, cannot be arbitrarily reversed.

A causal BN, in contrast, not only models the data distribution but also conforms to the principles of causality \cite{heckerman2013bayesian}. In the context of cause-effect relationship, intervening on the causes should render the effect independent of other factors. This introduces additional requirements for representing causality in a BN. In a causal BN, intervening on any subset of variables \( X_{I} \subseteq X \), denoted as \( do(X_{I}=x) \), results in a modified probability distribution \( P_{I}(X) \). This is computed by severing the edges from each variable in \( X_I \) to its parents and fixing their values as per the intervention:
\[
P_{I}(X) = \prod_{X_i\notin X_I} P(X_i \mid \textbf{Pa}_i^{\mathcal{G}})\quad \text{for all $X$ consistent with $x$} 
\]
This aspect of causal BNs allows for the modeling of interventions and causal inferences, distinguishing them from standard BNs.
It is important to note that in real-world scenarios, direct intervention data is often not available. As a result, observed data is typically employed to infer intervention characteristics and understand causal relationships.

\subsection{Learning Causal BNs}

This part introduces the task of two mainstream solutions of learning causal BNs, constraint- and score-based methods.
Formally, let $\mathbf{D} \in \mathbb{N}^{m \times n}$ represent the observational data, where $m$ denotes the number of observed samples and $n$ represents the number of observed variables, denoted as $X = \{X_1, X_2, \ldots, X_n\}$. Each $X_i$ in $\mathbf{D}$ takes discrete integer values in the range $[0, C_i)$. Given $\mathbf{D}$, the goal is to determine the causal DAG $\mathcal{G} = (X, E(\mathcal{G}))$, where $E(\mathcal{G})$ denotes the set of directed causal edges among the variables in $X$. The formal definitions are present as follows:

\begin{gather}
E(\mathcal{G}) \gets \{ X_i - X_j \mid X_i \not\!\perp\!\!\!\perp X_j \mid Y, \,\, \forall\, Y \subseteq X \setminus \{X_i, X_j\} \} \label{def_cons_CSL} \\
\max_{\mathcal{G}} \sigma(\mathcal{G}; \mathbf{D}) = \sum_{i=1}^{n} \mathcal{L}_{\sigma}(X_i \mid \textbf{Pa}_i^{\mathcal{G}}; \mathbf{D}) \,\, \text{s.t.} \,\, \mathcal{G} \in \mathrm{DAG} \label{def_score_CSL}
\end{gather}
Equations (\ref{def_cons_CSL}) and (\ref{def_score_CSL}) define the CSL task of constraint- and score-based methods, repectively. Constraint-based methods first determine the skeleton of the graph using undirected edges, \(X_i - X_j\), based on conditional independence tests. Subsequently, they orient some of these edges based on V-structure detection and DAG constraints~\cite{spirtes1991algorithm,strobl2018fast}.
Score-based methods employ a scoring function, $\sigma$, to evaluate how well a given causal DAG $\mathcal{G}$ represents the observed data $\mathbf{D}$. Typically, $\sigma$ can be decomposed into scores of local structures, $\mathcal{L}_{\sigma}(X_i \mid \textbf{Pa}_i^{\mathcal{G}}; \mathbf{D})$, which simplifies the search process~\cite{heckerman1995learning,neath2012bayesian}. The objective is to optimize these local scores by assigning appropriate parent nodes to each node, ensuring the resulting graph is a DAG. An alternative approach to searching the DAG space is the ordering-based search, which optimizes Equation (\ref{def_score_CSL}) under a given ordering $O$, inherently satisfying the DAG constraint ~\cite{yuan2011learning,trosser2021improved}. The best-scored DAG of the searched orderings is then selected as the output. 

The design of scoring functions is based on the posterior probability of the DAG given the data, which includes a component representing the prior probability of DAG structures. 
Due to this adaptability in accommodating the prior constraints on structures, the score-based method is chosen as the backbone CSL algorithm in our ILS-CSL framework.

\subsection{Prior Constraints on Structures}
\label{sec_prelimianry_prior}
Prior structural constraints play a pivotal role in improving the discovery of causal structures. The most prevalent among these constraints include ~\cite{li2018bayesian}:

\begin{itemize}
\item \textbf{Edge Existence}: Denoted as \(X_i\rightarrow X_j\) or, when forbidden, \(X_i \nrightarrow X_j\). This constraint dictates that the DAG should (or should not) contain the edge \(X_i\rightarrow X_j\).

\item \textbf{Ordering Constraint}: Represented as \(X_i \prec X_j\), it mandates that \(X_i\) should precede \(X_j\) in the variable ordering.

\item \textbf{Path Existence (Ancestral Constraint)}: Symbolized as \(X_i\leadsto X_j\), it requires the DAG to encompass the path \(X_i \leadsto X_j\).
\end{itemize}

\begin{algorithm}[!h]
 \caption{LLM supervised CSL}
\label{alg:llm_csl}
\begin{algorithmic}[1]
\Require Observed data, $\mathbf{D}$; Textual descriptions, $\mathbf{T}$
\Ensure Causal DAG, $\mathcal{G}$

\State Initialize the set of structural constraints, $\lambda\gets \{\}$
\Repeat
\State $\mathcal{G} \gets \underset{\mathcal{G}}{\arg\max}\,\sigma(\mathcal{G};\mathbf{D}), \,\, \text{s.t. }\mathcal{G}\in\text{DAG},\mathcal{G}\models\lambda$ \label{line:prior_CSL}
\For{$X_i\rightarrow X_j\in E(\mathcal{G})$}
 \State $c \gets$ LLM infers causality between $X_i$ and $X_j$ based on $\mathbf{T}$ \label{line:llm_infer}
 \If{$c$ is $X_i\leftarrow X_j$} \label{line:specify_s}
\State $\lambda \gets \lambda \cup \{X_j\rightarrow X_i\}$
 \EndIf
 \If{$c$ is $X_i\nleftrightarrow X_j$}
\State $\lambda \gets \lambda \cup \{X_i \nrightarrow X_j, X_j\nrightarrow X_i\}$
 \EndIf \label{line:specify_e}
\EndFor
\Until{no new constraints are added}
\State \Return $\mathcal{G}$
\end{algorithmic}
\end{algorithm}

Given the implication chain \(X_i\rightarrow X_j \Rightarrow X_i\leadsto X_j \Rightarrow X_i \prec X_j\), it is clear that the existence of an edge (direct causality) represents the most stringent structural constraint.
Correspondingly, its derivation necessitates a thorough examination of potential combinations of causality. 
Regrettably, as evidenced by the studies~\cite{kiciman2023causal,ban2023query,tu2023causal},
LLMs lack the ability to accurately specify direct causality, often confusing it with indirect causality or non-causal correlations. Please refer to Appendix \ref{app_sec_est_prec} for empirical estimation. 

Regarding the application of these prior constraints, there are two predominant methodologies: hard and soft approaches. The hard approach prioritizes adherence to prior constraints, followed by score optimization~\cite{de2007bayesian}.
Conversely, the soft approach strikes a balance between honoring prior constraints and the associated score costs~\cite{amirkhani2016exploiting}. This often involves adjusting the scoring function to \(\sigma(\mathcal{G};\mathbf{D})+b(\mathcal{G};\lambda)\), where a prior probability \(P_{\lambda}\) is assigned to structural constraints \(\lambda\). A constraint is only accepted if the bonus score, \(b\), compensates for the penalty in the DAG-data consistency score, \(\sigma\).

We implement both hard and soft approaches to incorporate structural constraints in this paper.

\section{Iterative LLM Supervised Causal Structure learning}
\label{sec_method}

Given the observed data, $\mathbf{D}$, and the descriptive texts on the investigated field and variables, $\mathbf{T}$, the LLM supervised causal structure learning is presented in Algorithm \ref{alg:llm_csl}.

Initially, a causal DAG $\mathcal{G}$ is learned from $\mathbf{D}$ with modular scoring function $\sigma,\mathcal{L}_{\sigma}$ (see Equation (\ref{def_score_CSL}) for definition), and search method $\mathcal{M}$.
Subsequently, we explicate the details on LLM supervision and how to constrain CSL accordingly.

\subsection{LLM Supervision}
For each directed edge $X_i\rightarrow X_j\in E(\mathcal{G})$, we prompt the used LLM to verify the causal statement that $X_i$ \texttt{causes} $X_j$ (Line \ref{line:llm_infer} in Algorithm \ref{alg:llm_csl}).
The prompt design for causal inference is inspired by the work~\cite{kiciman2023causal}, which employs choice-based queries to determine the orientation of pairwise variables with known causal relationships.
On this basis, we incorporate field-specific descriptions to provide context and introduce additional choices to accommodate uncertainties in causal existence and intricate causal mechanisms.
For a given edge $X_i\rightarrow X_j$ and associated textual descriptions $\mathbf{T}=\{t_f,t_i,t_j\}$, the LLM is prompted as:

\begin{flushleft}
\footnotesize \texttt{You are an expert on $t_f$. There 
are two factors:$X_i:t_i$,$X_j:t_j$.\\
Which cause-and-effect relationship is more likely 
for following causal statements for V1 and V2?\\
A.changing V1 causes a change in V2.\\
B.changing V2 causes a change in V1.\\
C.changes in V1 and in V2 are not correlated.\\
D.uncertain.\\
Provide your final answer within the tags <Answer>A/B/C/D</Answer>.\\ Analyze the statement:$X_i\,\,X_j$.}
\end{flushleft}

$t_f$ describes the investigated field, and $t_i,t_j$ describes $X_i,X_j$, respectively.
From the LLM's response to this prompt, we can obtain one of the answers: A, B, C, or D.

To specify constraints $\lambda$ (Lines \ref{line:specify_s}-\ref{line:specify_e} in Algorithm \ref{alg:llm_csl}),
if the answer is B (reversed), we specify the existence of $X_j\rightarrow X_i$.
If C (no causality), then we specify $X_i \nleftrightarrow X_j$ to forbid the existence of edge.
If D (uncertain) or A (correct), we do not specify constraints.
This is because specifying the existence of an edge already discovered from data does not often enhance the CSL and can inadvertently lead to errors.
For example, if the true structure is $X_i \leadsto X_j$ but not directly, $X_i\nrightarrow X_j$, LLM easily infers that $X_i$ \texttt{causes} $X_j$ due to its shortness in distinguishing indirect causality for the direct.
If we specify $X_i\rightarrow X_j$, an erroneous edge is introduced.

\subsection{Prior constraint-based CSL}

With the structural constraints $\lambda$ obtained from LLM supervision, we integrate them into the next iteration of CSL process (Line \ref{line:prior_CSL} in Algorithm \ref{alg:llm_csl}), with either hard or soft approach.
The process terminates if no new constraint is specified.

\paragraph{Hard approach} 
Firstly, the edge existence and forbidden constraints are used to specify the set of legal candidate parents, $C(i)$, and the set of variables always included in the parents, $K(i)$, of each variable $X_i$.
\begin{equation}
\begin{split}
C(i) &= X\setminus \{X_j\mid X_j\nrightarrow X_i \in \lambda\} \setminus \{X_i\}\\
    K(i) &= \{X_j\mid X_j\rightarrow X_i \in \lambda\} 
\end{split}
\end{equation}
With $K(i),C(i)$, we prune the space of local structures.
\begin{gather}
 L(X_i;\lambda) = \{P\mid K(i)\subseteq P \subseteq C(i) \} 
\end{gather}
The pruned space of local structures, $L(\cdot)$, is taken as input for the search method $\mathcal{M}$:
\begin{equation}
\begin{split}
    &\mathcal{M}: \max_{\textbf{Pa}_i^{\mathcal{G}}} \sum_{i}^{n} \mathcal{L}_{\sigma}(X_i\mid \textbf{Pa}_i^{\mathcal{G}};\mathbf{D})\\ 
    &\text{s.t.}\, \mathcal{G}\in \text{DAG},\, \textbf{Pa}_i^{\mathcal{G}}\in L(X_i;\lambda) 
\end{split}
\end{equation}
In comparison to the problem form without prior constraints, as presented in Equation (\ref{def_score_CSL}), the restriction of the candidate parent sets of each node, $\textbf{Pa}_i^{\mathcal{G}}\in L(X_i;\lambda)$, ensures that the output DAG absolutely satisfies every edge constraint, $\mathcal{G}\models \lambda$.

\paragraph{Soft approach} We adapt the scoring function to model the edge constraints as follows:
\begin{gather}
\sigma^{\prime}(\mathcal{G};\mathcal{D},\lambda) = \sum_{i}^{n} \mathcal{L}_{\sigma}(X_i\mid \textbf{Pa}_i^{\mathcal{G}};\mathbf{D}) +\mathcal{L}_b(X_i,\textbf{Pa}_i^{\mathcal{G}};\lambda) 
\label{eq_soft_score} \\
\begin{split}
&\mathcal{L}_b(X_i,\textbf{Pa}_i^{\mathcal{G}};\lambda) =\\
& \sum_{X_j\rightarrow X_i\in \lambda} \left( \mathbb{I}_{X_j\in \textbf{Pa}_i^{\mathcal{G}}} \log P_{\lambda}  +
 \mathbb{I}_{X_j \not\in \textbf{Pa}_i^{\mathcal{G}}} \log\left(1-P_{\lambda}\right) \right)  \\
 +&\sum_{X_j\nrightarrow X_i\in \lambda} \left( \mathbb{I}_{X_j\in \textbf{Pa}_i^{\mathcal{G}}} \log \left(1-P_{\lambda}\right)
 +\mathbb{I}_{X_j \not\in \textbf{Pa}_i^{\mathcal{G}}} \log P_{\lambda} \right)
\end{split}
\label{eq_soft_local}
\end{gather}
This formulation is grounded in the decomposability of edge constraints. A detailed derivation can be found in Section \ref{app_soft_score}. 
$\mathbb{I}_{\texttt{condition}}$ is the indicator function, which takes the value $1$ if the \texttt{condition} is true and $0$ otherwise.
$P_{\lambda}$ is the prior confidence, a hyper-parameter.
Then search method $M$ optimizes the modified score:
\begin{equation}
 \mathcal{M}: \max_{\mathcal{G}} \sum_{i}^{n} \mathcal{L}_{\sigma}(X_i\mid \textbf{Pa}_i^{\mathcal{G}};\mathbf{D})+\mathcal{L}_b(X_i,\textbf{Pa}_i^{\mathcal{G}};\lambda),\,\, \text{s.t.}\, \mathcal{G}\in \text{DAG}
\end{equation}
The bonus score, $\mathcal{L}_b$, favors DAGs that align more closely with the structural constraints.
Note that a constraint will not be satisfied if it excessively penalizes the score $\mathcal{L}_{\sigma}$.

To sum up, while the hard approach derives greater benefits from accurate constraints (at the risk of being more sensitive to errors), the soft approach might not always adhere to all correct constraints but offers a degree of resilience against potential inaccuracies.

\section{Analysis of Key Concerns}

Theoretically quantifying the impact of prior knowledge on learned causal structures is difficult, mainly due to the complex and unpredictable nature of data insufficiency and noise. Analyzing the disparity between data-implied causal structures and actual causal truths is intricate. Making strict assumptions for analytical purposes might not reflect real-world scenarios, potentially leading to theoretical conclusions with limited practical applicability.

Nevertheless, we can examine two primary aspects of prior knowledge under simple and general assumptions: 1) the ability of the applied prior knowledge to correct causal structures, and 2) the alignment of the quality of this derived prior knowledge with the actual causal structures. These aspects provide a more tangible and realistic assessment of the effectiveness of prior knowledge in causal discovery.

\subsection{Correction of Prior Independent Structures}

Causal discovery fundamentally seeks to uncover unknown causal mechanisms. The role of prior knowledge, representing known causality, extends beyond merely adjusting the final output; it should ideally enhance the accuracy of the reconstructed causal structures. A key question is whether a prior constraint can indirectly influence and correct edges that are not directly governed by this knowledge.

\begin{figure}[!h]
    \centering
    \includegraphics{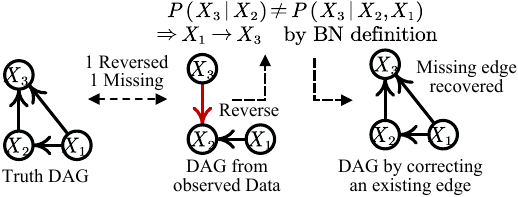}
    \caption{An example of recovering missing edges by reversing existing edges.}
    \label{fig_missing}
\end{figure}

In the context of ILS-CSL, this question becomes particularly relevant when examining the orientation and prohibition of learned edges: Do these constraints contribute to identifying missing edges? We explore this aspect, offering an illustrative example in Figure \ref{fig_missing}.
Due to limitations in real-world observational data, the probability distribution suggested by the data corresponds to a DAG with two errors: one reversed edge and one missing edge.

ILS-CSL supervises the existing edges and corrects the reversed edge \(X_3 \rightarrow X_2\). According to Bayesian Network principles, we have \(P(X_3 \mid \textbf{Pa}_3^{\mathcal{G}}) = P(X_3 \mid X_1, X_2)\). However, the observed data indicate that \(X_3\) and \(X_1\) are not independent when conditioned on \(X_2\), as per the current DAG structure. This inconsistency implies that the BN cannot accurately model the data distribution if \(X_2\) is the only parent of \(X_3\). Consequently, ILS-CSL identifies and reinstates the missing edge \(X_1 \rightarrow X_3\), thus refining the DAG to better align with the underlying data distribution.




Viewing this from the lens of knowledge-based causality, constraints derived from known causal relations can enhance the discovery of unknown causal mechanisms within data.
This highlights the invaluable role of prior knowledge in advancing causal discovery in uncharted fields.

\subsection{Estimation of Prior Error Counts}
\label{sec_theorem}

This section estimates and compares the number of erroneous constraints ILS-CSL against that stemming from a full inference on all pairwise variables, an intuitive strategy in the existing methods~\cite{ban2023query,vashishtha2023causal}.

We commence by defining five cases during LLM-based causality inference, along with their respective probabilities:

\begin{enumerate}
\item Extra Causality ($p_e$): Given a causal statement $(X_1,X_2)$, if the true causal DAG neither contains the path $X_1\leadsto X_2$ nor $X_2\leadsto X_1$, it's an instance of extra causality.

\item Reversed Causality ($p_r$): Given a causal statement $(X_1,X_2)$, if the true causal DAG contains the path $X_2\leadsto X_1$, it's an instance of reversed causality. 

\item Reversed Direct Causality ($p_{r}^d$): Given a causal statement $(X_1,X_2)$, if the true causal DAG has an edge \(X_2\rightarrow X_1\), it's an instance of extra causality.

\item Missing Direct Causality ($p_m^d$): If an edge $X_1\rightarrow X_2$ or $X_2\rightarrow X_1$ exist in the true causal DAG, but $X_1$ and $X_2$ are inferred to have no causal relationship, it's a instance of missing direct causality.

\item Correct Existing Causality ($p_c$): Given a causal statement $(X_1,X_2)$, if the path $X_1\leadsto X_2$ exists in the true causal DAG, it's a instance of correct existing causality.
\end{enumerate}
We assume that 1) the probability of these cases is identical when satisfying the corresponding structures, and 2) the truth DAG and learned DAG are both sparse.

Consider a causal DAG consisting of \( N \) nodes. Based on the sparsity assumption, the number of node pairs without connecting paths in the truth DAG is represented as \( \gamma_1 \binom{N}{2} \). In the learned causal DAG, there are \( \gamma_2 N \) edges. Of these edges, the proportion of correctly identified edges is denoted as \( z_1 \), the proportion of reversed edges as \( z_2 \), and the proportion of extra edges that do not exist in the true DAG as \( z_3 \).

The number of prior errors derived from full inference consists of two parts: the extra causality, \(p_e \gamma_1 \binom{N}{2}\), and the reversed causality, \(p_r(1-\gamma_1)\binom{N}{2}\).
Note that the missing causality will not harm the CSL since it does not produce any structural constraints in this context.
Then the total number of erroneous constraints is estimated as:
\begin{equation}
E_{\text{full}} = \left(p_e\gamma_1+p_r(1-\gamma_1)\right)\binom{N}{2}
\end{equation}

As for the prior errors within our framework, we consider the output DAG of CSL algorithms.
The erroneous constraints on the correctly discovered edges consist of the reversed and missing direct causality: $(p_r^d+p_m^d)z_1 \gamma_2 N$;
The erroneous constraints derived from inferring causality on erroneous edges consist of 1) missing direct causality on reversed edges, $p_m^d z_2 \gamma_2 N$, and 2) extra inferred direct causality on extra edges no more than $\left(p_r +p_c P_{R|E}\right) z_3 \gamma_2 N$, where $P_{R|E}$ is the probability where for an extra edge $X_1\rightarrow X_2$ in the learned DAG, a reversed path $X_2 \leadsto X_1$ exists in the ground truth.
Gathering all these, we derive the number prior errors:

\begin{equation}
E_{\text{ours}} \le \left((p_r^d +p^d_m)z_1 +p_m^d z_2+(p_r +p_c P_{R|E}) z_3 \right) \gamma_2 N
\end{equation}

We utilize eight real-world datasets, and GPT-4 as LLM to estimate $p$, and MIONBSx algorithm to estimate $\lambda,r,P_{R|E}$, see Section \ref{app_para_estimate} for details.
The results are:
\begin{equation}
 \begin{split}
&p_e\approx0.56,p_r\approx0.15,p_r^d \approx0.03,p_m^d\approx0.05\\
&p_c\approx0.75,\gamma_1\approx0.51,\gamma_2\approx1.09,z_1\approx0.88\\
&z_2\approx0.05,z_3\approx0.07,P_{R|E}\approx0.05
 \end{split}
\end{equation}
And then we have:
\begin{equation}
E_{\text{ours}} \approx 0.10N, E_{\text{full}} \approx 0.36\binom{N}{2}, \frac{E_{\text{ours}}}{E_{\text{full}}} \approx \frac{1}{1.8(N-1)}
\end{equation}
This indicates that, relative to full pairwise variable inference, ILS-CSL significantly reduces the number of erroneous constraints resulting from imperfect LLM inferences by approximately a factor of \(1.8(N-1)\). This reduction is particularly impactful when dealing with larger sets of variables.

\section{Supplementary Illustrations}

\subsection{Derivation of Prior-based Scoring}
\label{app_soft_score}
In this section, we derive the prior-based scoring function, as presented in Equations \eqref{eq_soft_score} and \eqref{eq_soft_local}, for the DAG \(\mathcal{G}(X,E(\mathcal{G}))\).
The prior constraints are denoted as $\lambda:$ \verb|<|$\mathbf{R},\boldsymbol{\Pi}$\verb|>|.
The set $\mathbf{R} =\{r_1,r_2,\cdots,r_m\}$ comprises edge variables on $m$ pairwise variables, where \( r_i\in \{\rightarrow, \nrightarrow \} \).
$\boldsymbol{\Pi}=\Pi_{i=1}^{m}P(r_i)$ is the associated probability distribution.

Beginning with the derivation of the scoring function without prior constraints, let \(\mathbf{D}\) be a complete multinomial observed data over variables \(X\). Utilizing the Bayesian Theorem, the probability of a network \( \mathcal{G} \) over \(X\) is expressed as:
\begin{equation}
 P(\mathcal{G}|\mathbf{D}) \propto P(\mathbf{D}|\mathcal{G}) \cdot P(\mathcal{G}) \notag
\end{equation}
Given that \(P(\mathbf{D})\) remains consistent across all DAGs, the score of a network is typically the logarithm of \( P(\mathcal{G}|\mathbf{D}) \), resulting in \( Sc(\mathcal{G}|\mathbf{D}) = Sc(\mathbf{D}|\mathcal{G}) +Sc(\mathcal{G}) \). Bayesian scoring methods, such as K2~\cite{cooper1992bayesian} and BDe, BDeu~\cite{heckerman1995learning}, aim to approximate the log-likelihood based on various assumptions. When priors are uniform, \( Sc(\mathcal{G}) \) can be disregarded during maximization. However, with the introduction of prior structural constraints, denoted as $\lambda$, this term gains significance.

Let's define $C$ as a configuration, representing a joint instantiation of values to edge variables $\mathbf{R}=\{r_1,r_2,...,r_m\}$. The probability for this configuration is \( J_C = P(\mathbf{R} = C|\boldsymbol{\Pi}) \). For a specific DAG $\mathcal{G}$, its configuration is represented as $C_\mathcal{G}$. Thus, we can express:

\begin{equation}
P(\mathcal{G}\mid \mathbf{D},\lambda) = \frac{P(\mathbf{D} \mid \mathcal{G}) \cdot P(\mathcal{G} \mid J)}{P(\mathbf{D} \mid J)}
\end{equation}
The above equation is derived from the understanding that, given the graph \(\mathcal{G}\), the data \(\mathbf{D}\) is independent of $J$. This is because $J$ offers no supplementary information about the data once the graph structure is known. The term $P(\mathbf{D} \mid J)$ serves as a normalizing constant, consistent across all DAGs. The term $P(\mathbf{D} \mid \mathcal{G})$ corresponds to the scoring function $Sc(\mathbf{D}\mid \mathcal{G})$ in the absence of prior constraints.
The scoring function can be expressed as:
\begin{equation}
Sc(\mathcal{G}\mid \mathbf{D},\lambda) = Sc(\mathbf{D}\mid \mathcal{G}) +Sc(\mathcal{G}\mid J)
\label{app_eq_prior_score}
\end{equation}
\begin{table*}[!h]
\setlength{\tabcolsep}{3pt}
\small
\caption{Accuracy and reversed ratio of the sampled pairwise variables on eight datasets.}
\begin{center}
 \begin{tabular}{c|cccccccc}
 \hline
Dataset  & Alarm& Asia & Insurance& Mildew & Child& Cancer & Water& Barley \\ \hline
Direct causality (Acc$_1$ /Rev$_1$) & 1.00 / 0.00 & 1.00 / 0.00 & 0.85 / 0.05 & 0.95 / 0.05 & 1.00 / 0.00 & 1.00 / 0.00 & 0.95 / 0.05 & 0.70 / 0.05 \\
Indirect causality (Acc$_2$ /Rev$_2$) & 0.65 / 0.15 & 1.00 / 0.00 & 0.95 / 0.05 & 1.00 / 0.00 & 0.50 / 0.40 & 1.00 / 0.00 & 0.50 / 0.50 & 0.30 / 0.30 \\
No causality (Acc$_3$) & 0.60  & 0.80  & 0.35  & 0.10  & 0.50  & 0.00  & 0.45  & 0.50  \\
Qualitative causality(Acc$_4$ / Rev$_4$) & 0.72 / 0.12 & 1.00 / 0.00 & 0.92 / 0.05 & 0.99 / 0.01 & 0.70 / 0.24 & 1.00 / 0.00 & 0.67 / 0.33 & 0.36 / 0.26\\\hline
\end{tabular}
\end{center}
\label{app_tab_esti}
\end{table*}
Here, $Sc(\mathbf{D}\mid \mathcal{G})$ represents the scoring function without prior constraints, denoted as $\sigma(\mathcal{G}\mid \mathbf{D})$. Meanwhile, $Sc(\mathcal{G}\mid J)$ pertains to the bonus score associated with prior constraints.
Shifting our focus to the prior factor $P(\mathcal{G} \mid J)$, we have:
\begin{equation}
\begin{aligned}
P(\mathcal{G} \mid J)= & P\left(\mathcal{G}, C_{\mathcal{G}} \mid J\right) = P\left(\mathcal{G} \mid J, C_{\mathcal{G}}\right) \cdot P\left(C_{\mathcal{G}} \mid J\right) \\
=& P\left(\mathcal{G} \mid C_{\mathcal{G}}\right) \cdot J_{C_{\mathcal{G}}}
\end{aligned}
\end{equation}
The first equation holds since $C_{\mathcal{G}}$ is inherently a function of $\mathcal{G}$. The term $P\left(\mathcal{G} \mid C_{\mathcal{G}}\right)$ denotes the likelihood of graph $\mathcal{G}$ when a specific configuration is present. In the absence of any other prior constraints, we assign an identical prior to all graphs sharing the same configuration. Let $N_C$ represent the count of DAGs over nodes $\mathcal{V}$ that have the configuration $C$. Thus, $P\left(\mathcal{G} \mid C_{\mathcal{G}}\right)=1 / N_{C_{\mathcal{G}}}$, leading to:
\begin{equation}
P(\mathcal{G} \mid J) = \frac{J_{C_{\mathcal{G}}}}{N_{C_{\mathcal{G}}}} \quad \text { and } \quad Sc(\mathcal{G} \mid J) = \log \left(\frac{J_{C_{\mathcal{G}}}}{N_{C_{\mathcal{G}}}}\right)
\end{equation}
Given that the count of edge variables (or edge constraints) remains consistent across all DAGs, $N_{C_{\mathcal{G}}}$ is also consistent for all DAGs. Therefore:
\begin{equation}
Sc(\mathcal{G}\mid J) = \log J_{C_{\mathcal{G}}} = \log P(\mathbf{R} = C_{\mathcal{G}}\mid \boldsymbol{\Pi}) = \sum_{r_i \in \mathbf{R}} \log P(r_i)
\end{equation}
Assuming $P(r_i) = P_{\lambda}$ when $\lambda$ indicates the presence of the corresponding edge, and $P(r_i) = 1 - P_{\lambda}$ when the edge's existence is negated, we deduce:
\begin{equation}
\begin{split}
&Sc(\mathcal{G}\mid J) =\\
& \sum_{X_j\rightarrow X_i \in \lambda} \mathbb{I}_{X_j\rightarrow X_i\in E(\mathcal{G})} \log P_{\lambda} +\mathbb{I}_{X_j\rightarrow X_i\not\in E(\mathcal{G})} \log (1-P_{\lambda}) +\\
& \sum_{X_j\nrightarrow X_i \in \lambda} \mathbb{I}_{X_j\rightarrow X_i\in E(\mathcal{G})} \log (1-P_{\lambda}) +\mathbb{I}_{X_j\rightarrow X_i\not\in E(\mathcal{G})} \log P_{\lambda}
\label{app_eq_local_prior_score}
\end{split}
\end{equation}

By integrating Equations (\ref{app_eq_prior_score}), (\ref{app_eq_local_prior_score}), and (\ref{def_score_CSL}), we derive the form of the local prior constraint-based scoring function, as depicted in Equations (\ref{eq_soft_score}) and (\ref{eq_soft_local}).

\subsection{Parameter Estimation in Section \ref{sec_theorem}}
\label{app_para_estimate}
This section presents the details on the estimation of parameters related to the quality of LLM based causal inference, $p_e,p_r,p_r^d,p_m^d,p_c$, structures of the true causal DAGs, $\gamma_1$, and structures of the learned causal DAGs, $\gamma_2,z_1,z_2,z_3,P_{R|E}$.

\paragraph{Quality of LLM causal inference}
We randomly sample three kinds of pairwise variables from the employed eight datasets in experiments:
\begin{enumerate}
 \item Direct edges: Sampling pairwise variables with direct edge $X_i\rightarrow X_j$ in the ground truth.
 \item Indirect path: Sampling pairwise variables without direct edge but with a directed path, $X_i \nrightarrow X_j,X_i \leadsto X_j$.
 \item Not connected: Sampling pairwise variables without any path, $X_i\not \leadsto X_j, X_j \not \leadsto X_i$.
\end{enumerate}
For each type, we sample 20 pairwise variables form each dataset, if more than 20 pairwise variables satisfying the condition exist in the causal DAG.
Or we use all the pairwise variables as samples.

Subsequently, we query GPT-4 the causality between each pairwise variables through the prompt in Section \ref{sec_method}.
The true answer of Types 1 and 2 is A, and that of Type 3 is C.
The accuracy of GPT-4 on different datasets on these samples together with the ratio of reversed inference (B for Types 1 and 2) are reported in Table \ref{app_tab_esti}.

Direct causality corresponds to direct edges, indirect causality to indirect paths, and no causality corresponds to not connected variables.
The accuracy and reversed ratio of LLM inference on them is obtained by experiments.
The qualitative causality corresponds the paths (including edges), whose accuracy is estimated by $\text{Acc}_4 = (\text{Acc}_1\times |E| +\text{Acc}_2\times |P|)/(|E|+|P|)$, where $|E|$ and $|P|$ represents the number of edges and indirect paths in the true causal DAG.

By weighted sum of the accuracy and reversed ratio, we obtain the estimation of them.
Then the probability of the five introduced error that GPT-4 makes are presented as follows:
\begin{enumerate}
 \item Extra causality: $p_e = 1 - \text{Acc}_3 = 0.56$
 \item Reversed causality: $p_r = \text{Rev}_4 = 0.15$
 \item Reversed direct causality: $p_r^d = \text{Rev}_1 = 0.03$
 \item Missing direct causality: $p_m^d = 1 - \text{Acc}_1 - \text{Rev}_1 = 0.05$
 \item Correct existing causality: $p_c = \text{Acc}_4 = 0.75$
\end{enumerate}
We see that the major errors of GPT-4 inference is sourced from the extra causality, which is because some intuitively correlated concepts may not generate real causal relations in an experiment with specific conditions.
And that is why we should refer to data for causal analysis.
However, GPT-4 is prone to infer correct causality on pairwise variables with direct causality, which is the base of our framework to efficiently improves the quality of learned causal DAGs.

\paragraph{Structural parameters}
The structural parameters is estimated by the average value of them on the eight datasets.
The ones related to the causal structure learning of each dataset is estimated by the average value of them on twelve segments of observed data, using MINOBSx search and BDeu score.
See the detailed results in Table \ref{app_tab_strutural_para}.

\begin{table}[!h]
\footnotesize
\setlength{\tabcolsep}{2pt}
\caption{The estimated structural paramters on eight datasets.}
\begin{center}
\begin{tabular}{c|cccccccc|c} \hline
Dataset & Alarm & Asia & Insurance & Mildew & Child & Cancer & Water & Barley & Avg. \\ \hline
$\gamma_1$ & 0.67  & 0.36 & 0.52 & 0.52& 0.66  & 0.20& 0.65  & 0.52 & 0.51 \\
$\gamma_2$ & 1.22  & 1.01 & 1.44 & 0.79& 1.09  & 0.55& 1.34  & 1.27 & 1.09 \\
$z_1$ & 0.96  & 0.88 & 0.91 & 0.87& 0.98  & 0.90& 0.67  & 0.84 & 0.88 \\
$z_2$ & 0.02  & 0.00 & 0.05 & 0.08& 0.00  & 0.07& 0.12  & 0.07 & 0.05 \\
$z_3$ & 0.02  & 0.12 & 0.04 & 0.05& 0.02  & 0.03& 0.21  & 0.09 & 0.07 \\
$P_{R|E}$  & 0.02  & 0.00 & 0.05 & 0.08& 0.00  & 0.10& 0.12  & 0.08 & 0.05 \\\hline
\end{tabular}  
\end{center}
\label{app_tab_strutural_para}
\end{table}

\begin{table*}[!h]
\normalsize
\setlength{\tabcolsep}{4pt}
\caption{The used datasets of causal DAGs.}
\begin{center}
\begin{tabular}{c|cccccccc}
\hline
Dataset& Cancer & Asia & Child & Alarm & Insurance & Water & Mildew & Barley \\ \hline
Variables& 5& 8& 20& 37& 27 & 32& 35 & 48 \\
Edges& 4& 8& 25& 46& 52 & 66& 46 & 84 \\
Parameters & 10 & 18 & 230 & 509 & 1008& 10083 & 540150 & 114005\\
Data size & 250 / 1000 & 250 / 1000 & 500 / 2000 & 1000 / 4000 & 500 / 2000 & 1000 / 4000 & 8000 / 32000 & 2000 / 8000 \\\hline
\end{tabular}
\end{center}
\label{tab_dataset}
\end{table*}

\begin{table*}[!h]
\footnotesize
\setlength{\tabcolsep}{4pt}
\caption{Scaled SHD$\downarrow$ comparison to data-based and LLM-driven CSL.}
\begin{center}
\begin{tabular}{l|ll|ll|ll|ll}
\hline \multicolumn{1}{c|}{Dataset}& \multicolumn{2}{c|}{Cancer}& \multicolumn{2}{c|}{Asia}& \multicolumn{2}{c|}{Child} & \multicolumn{2}{c}{Insurance} \\
 \multicolumn{1}{c|}{N}&  \multicolumn{1}{c}{250}&  \multicolumn{1}{c|}{1000} &  \multicolumn{1}{c}{250}&  \multicolumn{1}{c|}{1000} &  \multicolumn{1}{c}{500}&  \multicolumn{1}{c|}{2000} &  \multicolumn{1}{c}{500}&  \multicolumn{1}{c}{2000}\\ \hline
MINOBSx&0.75{\scriptsize ±0.22}&0.46{\scriptsize ±0.29}&0.52{\scriptsize ±0.32}&0.31{\scriptsize ±0.07}&0.38{\scriptsize ±0.08}&0.21{\scriptsize ±0.04}&0.46{\scriptsize ±0.05}&0.29{\scriptsize ±0.02}\\
+sepLLM-hard& 0.13\qquad \raisebox{0.5ex}{\scriptsize\textcolor{red}{-83\%}} & 0.00\qquad \raisebox{0.5ex}{\scriptsize\textcolor{red}{-100\%}}& 0.27\qquad \raisebox{0.5ex}{\scriptsize\textcolor{red}{-48\%}} & 0.04\qquad \raisebox{0.5ex}{\scriptsize\textcolor{red}{-87\%}}& 0.42\qquad \raisebox{0.5ex}{\scriptsize\textcolor{blue}{+11\%}} & 0.31\qquad \raisebox{0.5ex}{\scriptsize\textcolor{blue}{+48\%}} & 0.91\qquad \raisebox{0.5ex}{\scriptsize\textcolor{blue}{+98\%}} & 0.60\qquad \raisebox{0.5ex}{\scriptsize\textcolor{blue}{+107\%}}\\
+ILS-CSL-hard&0.50{\scriptsize ±0.22}\raisebox{0.5ex}{\scriptsize\textcolor{red}{-33\%}}&0.29{\scriptsize ±0.29}\raisebox{0.5ex}{\scriptsize\textcolor{red}{-37\%}}&0.42{\scriptsize ±0.37}\raisebox{0.5ex}{\scriptsize\textcolor{red}{-19\%}}&0.15{\scriptsize ±0.15}\raisebox{0.5ex}{\scriptsize\textcolor{red}{-52\%}}&0.25{\scriptsize ±0.06}\raisebox{0.5ex}{\scriptsize\textcolor{red}{-34\%}}&0.07{\scriptsize ±0.03}\raisebox{0.5ex}{\scriptsize\textcolor{red}{-67\%}}&0.42{\scriptsize ±0.03}\raisebox{0.5ex}{\scriptsize\textcolor{red}{-9\%}}&0.28{\scriptsize ±0.06}\raisebox{0.5ex}{\scriptsize\textcolor{red}{-3\%}}\\\hline
CaMML&0.75{\scriptsize ±0.00}&0.62{\scriptsize ±0.14}&0.58{\scriptsize ±0.29}&0.27{\scriptsize ±0.05}&0.25{\scriptsize ±0.03}&0.09{\scriptsize ±0.04}&0.69{\scriptsize ±0.04}&0.61{\scriptsize ±0.15}\\
+sepLLM-soft& 0.50\qquad \raisebox{0.5ex}{\scriptsize\textcolor{red}{-33\%}}& 0.33\qquad \raisebox{0.5ex}{\scriptsize\textcolor{red}{-47\%}} & 0.02\qquad \raisebox{0.5ex}{\scriptsize\textcolor{red}{-97\%}} & 0.00\qquad \raisebox{0.5ex}{\scriptsize\textcolor{red}{-100\%}} & 0.19\qquad \raisebox{0.5ex}{\scriptsize\textcolor{red}{-24\%}} & 0.04\qquad \raisebox{0.5ex}{\scriptsize\textcolor{red}{-56\%}} & 1.00\qquad \raisebox{0.5ex}{\scriptsize\textcolor{blue}{+45\%}} & 0.82\qquad \raisebox{0.5ex}{\scriptsize\textcolor{blue}{+34\%}}\\
+ILS-CSL-soft&0.75{\scriptsize ±0.00}\raisebox{0.5ex}{\scriptsize\textcolor{blue}{+0\%}}&0.33{\scriptsize ±0.20}\raisebox{0.5ex}{\scriptsize\textcolor{red}{-47\%}}&0.23{\scriptsize ±0.09}\raisebox{0.5ex}{\scriptsize\textcolor{red}{-60\%}}&0.15{\scriptsize ±0.18}\raisebox{0.5ex}{\scriptsize\textcolor{red}{-44\%}}&0.17{\scriptsize ±0.05}\raisebox{0.5ex}{\scriptsize\textcolor{red}{-32\%}}&0.04{\scriptsize ±0.00}\raisebox{0.5ex}{\scriptsize\textcolor{red}{-56\%}}&0.47{\scriptsize ±0.04}\raisebox{0.5ex}{\scriptsize\textcolor{red}{-32\%}}&0.47{\scriptsize ±0.11}\raisebox{0.5ex}{\scriptsize\textcolor{red}{-23\%}}\\\hline \hline
\multicolumn{1}{c|}{Dataset}& \multicolumn{2}{c|}{Alarm} & \multicolumn{2}{c|}{Mildew} & \multicolumn{2}{c|}{Water} & \multicolumn{2}{c}{Barley}\\
\multicolumn{1}{c|}{N}& \multicolumn{1}{c}{1000} & \multicolumn{1}{c|}{4000} & \multicolumn{1}{c}{8000} & \multicolumn{1}{c|}{32000} & \multicolumn{1}{c}{1000} & \multicolumn{1}{c|}{4000} & \multicolumn{1}{c}{2000} & \multicolumn{1}{c}{8000} \\ \hline
MINOBSx&0.21{\scriptsize ±0.06}&0.14{\scriptsize ±0.04}&0.50{\scriptsize ±0.02}&0.46{\scriptsize ±0.05}&0.77{\scriptsize ±0.07}&0.61{\scriptsize ±0.04}&0.56{\scriptsize ±0.04}&0.40{\scriptsize ±0.03}\\
+sepLLM-hard& 0.27\qquad\raisebox{0.5ex}{\scriptsize\textcolor{blue}{+29\%}} & 0.19\qquad \raisebox{0.5ex}{\scriptsize\textcolor{blue}{+36\%}}& 0.88\qquad \raisebox{0.5ex}{\scriptsize\textcolor{blue}{+76\%}} & 0.47\qquad \raisebox{0.5ex}{\scriptsize\textcolor{blue}{+2\%}}& 1.01\qquad \raisebox{0.5ex}{\scriptsize\textcolor{blue}{+31\%}} & 0.84\qquad \raisebox{0.5ex}{\scriptsize\textcolor{blue}{+38\%}} & 0.62\qquad \raisebox{0.5ex}{\scriptsize\textcolor{blue}{+11\%}} & 0.65\qquad \raisebox{0.5ex}{\scriptsize\textcolor{blue}{+62\%}}\\
+ILS-CSL-hard&0.09{\scriptsize ±0.03}\raisebox{0.5ex}{\scriptsize\textcolor{red}{-57\%}}&0.08{\scriptsize ±0.02}\raisebox{0.5ex}{\scriptsize\textcolor{red}{-43\%}}&0.43{\scriptsize ±0.00}\raisebox{0.5ex}{\scriptsize\textcolor{red}{-14\%}}&0.33{\scriptsize ±0.18}\raisebox{0.5ex}{\scriptsize\textcolor{red}{-28\%}}&0.68{\scriptsize ±0.05}\raisebox{0.5ex}{\scriptsize\textcolor{red}{-12\%}}&0.56{\scriptsize ±0.02}\raisebox{0.5ex}{\scriptsize\textcolor{red}{-8\%}}&0.54{\scriptsize ±0.02}\raisebox{0.5ex}{\scriptsize\textcolor{red}{-4\%}}&0.38{\scriptsize ±0.02}\raisebox{0.5ex}{\scriptsize\textcolor{red}{-5\%}}\\\hline
CaMML&0.24{\scriptsize ±0.05}&0.18{\scriptsize ±0.06}&1.20{\scriptsize ±0.10}&1.30{\scriptsize ±0.12}&0.88{\scriptsize ±0.08}&0.81{\scriptsize ±0.04}&0.96{\scriptsize ±0.07}&0.96{\scriptsize ±0.10}\\
+sepLLM-soft& 0.13\qquad\raisebox{0.5ex}{\scriptsize\textcolor{red}{-46\%}} & 0.07\qquad \raisebox{0.5ex}{\scriptsize\textcolor{red}{-61\%}}& 1.07\qquad \raisebox{0.5ex}{\scriptsize\textcolor{red}{-11\%}} & 1.30\qquad \raisebox{0.5ex}{\scriptsize\textcolor{blue}{+0\%}}& 0.89\qquad \raisebox{0.5ex}{\scriptsize\textcolor{blue}{+1\%}} & 0.73\qquad \raisebox{0.5ex}{\scriptsize\textcolor{red}{-10\%}} & 0.98\qquad \raisebox{0.5ex}{\scriptsize\textcolor{blue}{+2\%}} & 0.98\qquad \raisebox{0.5ex}{\scriptsize\textcolor{blue}{+2\%}}\\
+ILS-CSL-soft&0.08{\scriptsize ±0.01}\raisebox{0.5ex}{\scriptsize\textcolor{red}{-67\%}}&0.06{\scriptsize ±0.01}\raisebox{0.5ex}{\scriptsize\textcolor{red}{-67\%}}&1.01{\scriptsize ±0.07}\raisebox{0.5ex}{\scriptsize\textcolor{red}{-16\%}}&1.26{\scriptsize ±0.05}\raisebox{0.5ex}{\scriptsize\textcolor{red}{-3\%}}&0.70{\scriptsize ±0.02}\raisebox{0.5ex}{\scriptsize\textcolor{red}{-20\%}}&0.63{\scriptsize ±0.04}\raisebox{0.5ex}{\scriptsize\textcolor{red}{-22\%}}&0.90{\scriptsize ±0.06}\raisebox{0.5ex}{\scriptsize\textcolor{red}{-6\%}}&0.83{\scriptsize ±0.06}\raisebox{0.5ex}{\scriptsize\textcolor{red}{-14\%}}\\\hline
\end{tabular}
\end{center}
\begin{flushleft}
 \scriptsize The suffixes `-hard' and '-soft' represent the approach to apply the LLM inferred prior constraints.
 The performances of sepLLM method are obtained from the work~\cite{ban2023query}.
\end{flushleft}
\label{tab_cmp_ban}
\end{table*}

\section{Experiments}

We conduct experiments to address the research questions:

\noindent\textbf{RQ1:} Can ILS-CSL enhance data-based CSL baselines and outperform the existing LLM-driven CSL method?

\noindent\textbf{RQ2:} Across diverse backbone algorithms, can ILS-CSL consistently improve the quality of causal structures?
which of the soft and hard constraint is better?

\noindent\textbf{RQ3:} Is ILS-CSL resistant to imperfect LLM causal inferences, and capable to derive accurate prior? Why?

\noindent\textbf{RQ4:} How does the process, where LLM supervises causal discovery, unfold in detail?

All the datasets, codes, and supplementary results can be accessed in the external repository\footnote{\url{https://github.com/tyMadara/ILS-CSL}}.

\subsection{Datasets and Baselines}
To address RQ1, we employ the eight real-world datasets of causal DAGs from the Bayesian Network Repository\footnote{\url{https://www.bnlearn.com/bnrepository/}} as used in the comparative study~\cite{ban2023query}.
Dataset specifics are provided in Table \ref{tab_dataset}. 
For backbone CSL algorithms, we adopt the same MINOBSx (BDeu score)~\cite{li2018bayesian} and CaMML (MML score)~\cite{CaMML} algorithms, and utilize the same setting of prior probability for CaMML, 0.99999.
For supervision on CSL, we utilize GPT-4-WEB\footnote{\url{https://chat.openai.com/}}.
For RQ2, the used baselines comprise a combination of popular scoring functions, namely BIC and BDeu score~\cite{heckerman1995learning}, and search algorithms, including HC~\cite{gamez2011learning} and MINOBSx~\cite{lee2017metaheuristics}.

\subsection{Observed Data and Evaluation Metric}
We utilize a collection of observed data sourced from a public repository\footnote{\url{https://github.com/andrewli77/MINOBS-anc/tree/master/data/csv}}. This data, generated based on the eight causal DAGs, is provided by Li and Beek~\cite{li2018bayesian}, and used in the comparative work \cite{ban2023query}. The repository offers datasets in two distinct sample sizes for each DAG, as detailed in Table \ref{tab_dataset}. For every sample size, six distinct data segments are available.

To assess the quality of the learned causal structures, we primarily employ the scaled Structural Hamming Distance (SHD)~\cite{scutari2019learns}. This metric is defined as the SHD normalized by the total number of edges in the true causal DAG. 

\begin{table}[!h]
\small
\setlength{\tabcolsep}{3pt}
\caption{Ranking of methods in Table \ref{tab_cmp_ban}.}
\centering
\begin{tabular}{cc||cc|cc}
\hline
\multicolumn{2}{c||}{Data-based CSL} & \multicolumn{2}{c|}{SepLLM} & \multicolumn{2}{c}{ILS-CSL} \\ \hline
MINOBSx  & CaMML  & MINOBSx& CaMML& MINOBSx& CaMML\\ \hline
3.6 & 4.8 & 4.0  & 3.9& \textbf{1.9}  & \underline{2.9} \\ \hline
\end{tabular}
\label{tab_app_rank_1}
\end{table}

\subsection{Comparison Experiments (RQ1)}

We compare the performance of MINOBSx (BDeu) and CaMML that are used in the separate LLM prior-driven CSL approach proposed by \cite{ban2023query}, referred to as sepLLM, and our proposed framework, termed ILS-CSL. 
This comparison is conducted using all the introduced observed data across eight datasets. 
The results, presented in terms of scaled SHD (where a lower value is preferable), are detailed in Table \ref{tab_cmp_ban}.
The difference between scaled SHD of data-based ($\Delta_{\text{data}}$) and LLM-driven ($\Delta_{\text{LLM}}$) CSL is also reported, by calculating $(\Delta_{\text{LLM}}-\Delta_{\text{data}})/\Delta_{\text{data}}$.
The Friedman ranking of the methods and more is reported in Table \ref{tab_app_rank_1}.

\begin{table*}[!h]
\small
\setlength{\tabcolsep}{4pt}
    \caption{Scaled SHD$\downarrow$ enhancement on data-based CSL with different scores, search algorithms and approaches to apply prior constraints, by the proposed framework.}
    \centering
    \begin{tabular}{l|ll|ll|ll|ll}
    \hline \multicolumn{1}{c|}{Dataset}& \multicolumn{2}{c|}{Cancer}& \multicolumn{2}{c|}{Asia}& \multicolumn{2}{c|}{Child} & \multicolumn{2}{c}{Insurance} \\
     \multicolumn{1}{c|}{N}&  \multicolumn{1}{c}{250}&  \multicolumn{1}{c|}{1000} &  \multicolumn{1}{c}{250}&  \multicolumn{1}{c|}{1000} &  \multicolumn{1}{c}{500}&  \multicolumn{1}{c|}{2000} &  \multicolumn{1}{c}{500}&  \multicolumn{1}{c}{2000}\\ \hline
    HC-BDeu&0.58{\scriptsize ±0.13}&0.33{\scriptsize ±0.26}&0.56{\scriptsize ±0.27}&0.23{\scriptsize ±0.17}&0.57{\scriptsize ±0.12}&0.49{\scriptsize ±0.18}&0.69{\scriptsize ±0.06}&0.68{\scriptsize ±0.09}\\
    +ILS-CSL-hard&0.50{\scriptsize ±0.22}\raisebox{0.5ex}{\scriptsize\textcolor{red}{-14\%}}&0.29{\scriptsize ±0.29}\raisebox{0.5ex}{\scriptsize\textcolor{red}{-12\%}}&0.46{\scriptsize ±0.33}\raisebox{0.5ex}{\scriptsize\textcolor{red}{-18\%}}&0.15{\scriptsize ±0.15}\raisebox{0.5ex}{\scriptsize\textcolor{red}{-35\%}}&0.24{\scriptsize ±0.07}\raisebox{0.5ex}{\scriptsize\textcolor{red}{-58\%}}&0.10{\scriptsize ±0.02}\raisebox{0.5ex}{\scriptsize\textcolor{red}{-80\%}}&0.45{\scriptsize ±0.06}\raisebox{0.5ex}{\scriptsize\textcolor{red}{-35\%}}&0.34{\scriptsize ±0.04}\raisebox{0.5ex}{\scriptsize\textcolor{red}{-50\%}}\\
    +ILS-CSL-soft&0.50{\scriptsize ±0.22}\raisebox{0.5ex}{\scriptsize\textcolor{red}{-14\%}}&0.29{\scriptsize ±0.29}\raisebox{0.5ex}{\scriptsize\textcolor{red}{-12\%}}&0.44{\scriptsize ±0.30}\raisebox{0.5ex}{\scriptsize\textcolor{red}{-21\%}}&0.15{\scriptsize ±0.15}\raisebox{0.5ex}{\scriptsize\textcolor{red}{-35\%}}&0.26{\scriptsize ±0.06}\raisebox{0.5ex}{\scriptsize\textcolor{red}{-54\%}}&0.11{\scriptsize ±0.03}\raisebox{0.5ex}{\scriptsize\textcolor{red}{-78\%}}&0.50{\scriptsize ±0.08}\raisebox{0.5ex}{\scriptsize\textcolor{red}{-28\%}}&0.35{\scriptsize ±0.04}\raisebox{0.5ex}{\scriptsize\textcolor{red}{-49\%}}\\\hline
    MINOBSx-BDeu&0.75{\scriptsize ±0.22}&0.46{\scriptsize ±0.29}&0.52{\scriptsize ±0.32}&0.31{\scriptsize ±0.07}&0.38{\scriptsize ±0.08}&0.21{\scriptsize ±0.04}&0.46{\scriptsize ±0.05}&0.29{\scriptsize ±0.02}\\
    +ILS-CSL-hard&0.50{\scriptsize ±0.22}\raisebox{0.5ex}{\scriptsize\textcolor{red}{-33\%}}&0.29{\scriptsize ±0.29}\raisebox{0.5ex}{\scriptsize\textcolor{red}{-37\%}}&0.42{\scriptsize ±0.37}\raisebox{0.5ex}{\scriptsize\textcolor{red}{-19\%}}&0.15{\scriptsize ±0.15}\raisebox{0.5ex}{\scriptsize\textcolor{red}{-52\%}}&0.25{\scriptsize ±0.06}\raisebox{0.5ex}{\scriptsize\textcolor{red}{-34\%}}&0.07{\scriptsize ±0.03}\raisebox{0.5ex}{\scriptsize\textcolor{red}{-67\%}}&0.42{\scriptsize ±0.03}\raisebox{0.5ex}{\scriptsize\textcolor{red}{-9\%}}&0.28{\scriptsize ±0.06}\raisebox{0.5ex}{\scriptsize\textcolor{red}{-3\%}}\\
    +ILS-CSL-soft&0.50{\scriptsize ±0.22}\raisebox{0.5ex}{\scriptsize\textcolor{red}{-33\%}}&0.29{\scriptsize ±0.29}\raisebox{0.5ex}{\scriptsize\textcolor{red}{-37\%}}&0.42{\scriptsize ±0.37}\raisebox{0.5ex}{\scriptsize\textcolor{red}{-19\%}}&0.15{\scriptsize ±0.15}\raisebox{0.5ex}{\scriptsize\textcolor{red}{-52\%}}&0.25{\scriptsize ±0.04}\raisebox{0.5ex}{\scriptsize\textcolor{red}{-34\%}}&0.08{\scriptsize ±0.04}\raisebox{0.5ex}{\scriptsize\textcolor{red}{-62\%}}&0.41{\scriptsize ±0.03}\raisebox{0.5ex}{\scriptsize\textcolor{red}{-11\%}}&0.26{\scriptsize ±0.04}\raisebox{0.5ex}{\scriptsize\textcolor{red}{-10\%}}\\\hline
    HC-BIC&0.92{\scriptsize ±0.29}&0.62{\scriptsize ±0.34}&0.48{\scriptsize ±0.36}&0.31{\scriptsize ±0.29}&0.53{\scriptsize ±0.07}&0.38{\scriptsize ±0.16}&0.76{\scriptsize ±0.05}&0.72{\scriptsize ±0.06}\\
    +ILS-CSL-hard&0.92{\scriptsize ±0.29}\raisebox{0.5ex}{\scriptsize\textcolor{blue}{+0\%}}&0.42{\scriptsize ±0.34}\raisebox{0.5ex}{\scriptsize\textcolor{red}{-32\%}}&0.33{\scriptsize ±0.25}\raisebox{0.5ex}{\scriptsize\textcolor{red}{-31\%}}&0.19{\scriptsize ±0.17}\raisebox{0.5ex}{\scriptsize\textcolor{red}{-39\%}}&0.26{\scriptsize ±0.07}\raisebox{0.5ex}{\scriptsize\textcolor{red}{-51\%}}&0.07{\scriptsize ±0.03}\raisebox{0.5ex}{\scriptsize\textcolor{red}{-82\%}}&0.60{\scriptsize ±0.03}\raisebox{0.5ex}{\scriptsize\textcolor{red}{-21\%}}&0.41{\scriptsize ±0.03}\raisebox{0.5ex}{\scriptsize\textcolor{red}{-43\%}}\\
    +ILS-CSL-soft&0.92{\scriptsize ±0.29}\raisebox{0.5ex}{\scriptsize\textcolor{blue}{+0\%}}&0.42{\scriptsize ±0.34}\raisebox{0.5ex}{\scriptsize\textcolor{red}{-32\%}}&0.35{\scriptsize ±0.26}\raisebox{0.5ex}{\scriptsize\textcolor{red}{-27\%}}&0.21{\scriptsize ±0.19}\raisebox{0.5ex}{\scriptsize\textcolor{red}{-32\%}}&0.27{\scriptsize ±0.08}\raisebox{0.5ex}{\scriptsize\textcolor{red}{-49\%}}&0.07{\scriptsize ±0.05}\raisebox{0.5ex}{\scriptsize\textcolor{red}{-82\%}}&0.62{\scriptsize ±0.06}\raisebox{0.5ex}{\scriptsize\textcolor{red}{-18\%}}&0.42{\scriptsize ±0.03}\raisebox{0.5ex}{\scriptsize\textcolor{red}{-42\%}}\\\hline
    MINOBSx-BIC&1.00{\scriptsize ±0.25}&0.62{\scriptsize ±0.21}&0.46{\scriptsize ±0.23}&0.27{\scriptsize ±0.05}&0.34{\scriptsize ±0.06}&0.18{\scriptsize ±0.04}&0.62{\scriptsize ±0.05}&0.55{\scriptsize ±0.05}\\
    +ILS-CSL-hard&0.92{\scriptsize ±0.29}\raisebox{0.5ex}{\scriptsize\textcolor{red}{-8\%}}&0.38{\scriptsize ±0.26}\raisebox{0.5ex}{\scriptsize\textcolor{red}{-39\%}}&0.42{\scriptsize ±0.40}\raisebox{0.5ex}{\scriptsize\textcolor{red}{-9\%}}&0.12{\scriptsize ±0.08}\raisebox{0.5ex}{\scriptsize\textcolor{red}{-56\%}}&0.24{\scriptsize ±0.08}\raisebox{0.5ex}{\scriptsize\textcolor{red}{-29\%}}&0.06{\scriptsize ±0.02}\raisebox{0.5ex}{\scriptsize\textcolor{red}{-67\%}}&0.55{\scriptsize ±0.03}\raisebox{0.5ex}{\scriptsize\textcolor{red}{-11\%}}&0.39{\scriptsize ±0.08}\raisebox{0.5ex}{\scriptsize\textcolor{red}{-29\%}}\\
    +ILS-CSL-soft&0.92{\scriptsize ±0.29}\raisebox{0.5ex}{\scriptsize\textcolor{red}{-8\%}}&0.38{\scriptsize ±0.26}\raisebox{0.5ex}{\scriptsize\textcolor{red}{-39\%}}&0.35{\scriptsize ±0.26}\raisebox{0.5ex}{\scriptsize\textcolor{red}{-24\%}}&0.15{\scriptsize ±0.12}\raisebox{0.5ex}{\scriptsize\textcolor{red}{-44\%}}&0.25{\scriptsize ±0.05}\raisebox{0.5ex}{\scriptsize\textcolor{red}{-26\%}}&0.06{\scriptsize ±0.02}\raisebox{0.5ex}{\scriptsize\textcolor{red}{-67\%}}&0.55{\scriptsize ±0.03}\raisebox{0.5ex}{\scriptsize\textcolor{red}{-11\%}}&0.41{\scriptsize ±0.09}\raisebox{0.5ex}{\scriptsize\textcolor{red}{-25\%}}\\\hline
    \hline
    \multicolumn{1}{c|}{Dataset}& \multicolumn{2}{c|}{Alarm} & \multicolumn{2}{c|}{Mildew} & \multicolumn{2}{c|}{Water} & \multicolumn{2}{c}{Barley}\\
    \multicolumn{1}{c|}{N}& \multicolumn{1}{c}{1000} & \multicolumn{1}{c|}{4000} & \multicolumn{1}{c}{8000} & \multicolumn{1}{c|}{32000} & \multicolumn{1}{c}{1000} & \multicolumn{1}{c|}{4000} & \multicolumn{1}{c}{2000} & \multicolumn{1}{c}{8000} \\ \hline
    HC-BDeu&0.65{\scriptsize ±0.12}&0.64{\scriptsize ±0.09}&0.79{\scriptsize ±0.11}&0.99{\scriptsize ±0.07}&0.76{\scriptsize ±0.07}&0.64{\scriptsize ±0.08}&0.80{\scriptsize ±0.06}&0.65{\scriptsize ±0.06}\\
    +ILS-CSL-hard&0.12{\scriptsize ±0.02}\raisebox{0.5ex}{\scriptsize\textcolor{red}{-82\%}}&0.08{\scriptsize ±0.01}\raisebox{0.5ex}{\scriptsize\textcolor{red}{-88\%}}&0.46{\scriptsize ±0.01}\raisebox{0.5ex}{\scriptsize\textcolor{red}{-42\%}}&0.22{\scriptsize ±0.02}\raisebox{0.5ex}{\scriptsize\textcolor{red}{-78\%}}&0.64{\scriptsize ±0.02}\raisebox{0.5ex}{\scriptsize\textcolor{red}{-16\%}}&0.55{\scriptsize ±0.03}\raisebox{0.5ex}{\scriptsize\textcolor{red}{-14\%}}&0.69{\scriptsize ±0.06}\raisebox{0.5ex}{\scriptsize\textcolor{red}{-14\%}}&0.57{\scriptsize ±0.06}\raisebox{0.5ex}{\scriptsize\textcolor{red}{-12\%}}\\
    +ILS-CSL-soft&0.30{\scriptsize ±0.05}\raisebox{0.5ex}{\scriptsize\textcolor{red}{-54\%}}&0.25{\scriptsize ±0.06}\raisebox{0.5ex}{\scriptsize\textcolor{red}{-61\%}}&0.43{\scriptsize ±0.00}\raisebox{0.5ex}{\scriptsize\textcolor{red}{-46\%}}&0.47{\scriptsize ±0.04}\raisebox{0.5ex}{\scriptsize\textcolor{red}{-53\%}}&0.64{\scriptsize ±0.01}\raisebox{0.5ex}{\scriptsize\textcolor{red}{-16\%}}&0.56{\scriptsize ±0.03}\raisebox{0.5ex}{\scriptsize\textcolor{red}{-12\%}}&0.76{\scriptsize ±0.04}\raisebox{0.5ex}{\scriptsize\textcolor{red}{-5\%}}&0.62{\scriptsize ±0.03}\raisebox{0.5ex}{\scriptsize\textcolor{red}{-5\%}}\\\hline
    MINOBSx-BDeu&0.21{\scriptsize ±0.06}&0.14{\scriptsize ±0.04}&0.50{\scriptsize ±0.02}&0.46{\scriptsize ±0.05}&0.77{\scriptsize ±0.07}&0.61{\scriptsize ±0.04}&0.56{\scriptsize ±0.04}&0.40{\scriptsize ±0.03}\\
    +ILS-CSL-hard&0.09{\scriptsize ±0.03}\raisebox{0.5ex}{\scriptsize\textcolor{red}{-57\%}}&0.08{\scriptsize ±0.02}\raisebox{0.5ex}{\scriptsize\textcolor{red}{-43\%}}&0.43{\scriptsize ±0.00}\raisebox{0.5ex}{\scriptsize\textcolor{red}{-14\%}}&0.33{\scriptsize ±0.18}\raisebox{0.5ex}{\scriptsize\textcolor{red}{-28\%}}&0.68{\scriptsize ±0.05}\raisebox{0.5ex}{\scriptsize\textcolor{red}{-12\%}}&0.56{\scriptsize ±0.02}\raisebox{0.5ex}{\scriptsize\textcolor{red}{-8\%}}&0.54{\scriptsize ±0.02}\raisebox{0.5ex}{\scriptsize\textcolor{red}{-4\%}}&0.38{\scriptsize ±0.02}\raisebox{0.5ex}{\scriptsize\textcolor{red}{-5\%}}\\
    +ILS-CSL-soft&0.09{\scriptsize ±0.02}\raisebox{0.5ex}{\scriptsize\textcolor{red}{-57\%}}&0.07{\scriptsize ±0.01}\raisebox{0.5ex}{\scriptsize\textcolor{red}{-50\%}}&0.47{\scriptsize ±0.01}\raisebox{0.5ex}{\scriptsize\textcolor{red}{-6\%}}&0.37{\scriptsize ±0.02}\raisebox{0.5ex}{\scriptsize\textcolor{red}{-20\%}}&0.68{\scriptsize ±0.04}\raisebox{0.5ex}{\scriptsize\textcolor{red}{-12\%}}&0.56{\scriptsize ±0.02}\raisebox{0.5ex}{\scriptsize\textcolor{red}{-8\%}}&0.55{\scriptsize ±0.03}\raisebox{0.5ex}{\scriptsize\textcolor{red}{-2\%}}&0.38{\scriptsize ±0.02}\raisebox{0.5ex}{\scriptsize\textcolor{red}{-5\%}}\\\hline
    HC-BIC&0.68{\scriptsize ±0.05}&0.59{\scriptsize ±0.10}&0.90{\scriptsize ±0.06}&0.91{\scriptsize ±0.13}&0.76{\scriptsize ±0.04}&0.70{\scriptsize ±0.03}&0.87{\scriptsize ±0.05}&0.80{\scriptsize ±0.08}\\
    +ILS-CSL-hard&0.22{\scriptsize ±0.04}\raisebox{0.5ex}{\scriptsize\textcolor{red}{-68\%}}&0.12{\scriptsize ±0.04}\raisebox{0.5ex}{\scriptsize\textcolor{red}{-80\%}}&0.58{\scriptsize ±0.01}\raisebox{0.5ex}{\scriptsize\textcolor{red}{-36\%}}&0.46{\scriptsize ±0.04}\raisebox{0.5ex}{\scriptsize\textcolor{red}{-49\%}}&0.69{\scriptsize ±0.02}\raisebox{0.5ex}{\scriptsize\textcolor{red}{-9\%}}&0.61{\scriptsize ±0.03}\raisebox{0.5ex}{\scriptsize\textcolor{red}{-13\%}}&0.76{\scriptsize ±0.02}\raisebox{0.5ex}{\scriptsize\textcolor{red}{-13\%}}&0.69{\scriptsize ±0.06}\raisebox{0.5ex}{\scriptsize\textcolor{red}{-14\%}}\\
    +ILS-CSL-soft&0.41{\scriptsize ±0.04}\raisebox{0.5ex}{\scriptsize\textcolor{red}{-40\%}}&0.35{\scriptsize ±0.11}\raisebox{0.5ex}{\scriptsize\textcolor{red}{-41\%}}&0.71{\scriptsize ±0.01}\raisebox{0.5ex}{\scriptsize\textcolor{red}{-21\%}}&0.57{\scriptsize ±0.02}\raisebox{0.5ex}{\scriptsize\textcolor{red}{-37\%}}&0.69{\scriptsize ±0.02}\raisebox{0.5ex}{\scriptsize\textcolor{red}{-9\%}}&0.61{\scriptsize ±0.03}\raisebox{0.5ex}{\scriptsize\textcolor{red}{-13\%}}&0.82{\scriptsize ±0.04}\raisebox{0.5ex}{\scriptsize\textcolor{red}{-6\%}}&0.74{\scriptsize ±0.09}\raisebox{0.5ex}{\scriptsize\textcolor{red}{-8\%}}\\\hline
    MINOBSx-BIC&0.32{\scriptsize ±0.08}&0.15{\scriptsize ±0.04}&0.74{\scriptsize ±0.01}&0.73{\scriptsize ±0.09}&0.82{\scriptsize ±0.03}&0.77{\scriptsize ±0.03}&0.79{\scriptsize ±0.04}&0.58{\scriptsize ±0.03}\\
    +ILS-CSL-hard&0.16{\scriptsize ±0.07}\raisebox{0.5ex}{\scriptsize\textcolor{red}{-50\%}}&0.09{\scriptsize ±0.03}\raisebox{0.5ex}{\scriptsize\textcolor{red}{-40\%}}&0.58{\scriptsize ±0.01}\raisebox{0.5ex}{\scriptsize\textcolor{red}{-22\%}}&0.45{\scriptsize ±0.03}\raisebox{0.5ex}{\scriptsize\textcolor{red}{-38\%}}&0.69{\scriptsize ±0.03}\raisebox{0.5ex}{\scriptsize\textcolor{red}{-16\%}}&0.62{\scriptsize ±0.01}\raisebox{0.5ex}{\scriptsize\textcolor{red}{-19\%}}&0.73{\scriptsize ±0.03}\raisebox{0.5ex}{\scriptsize\textcolor{red}{-8\%}}&0.55{\scriptsize ±0.03}\raisebox{0.5ex}{\scriptsize\textcolor{red}{-5\%}}\\
    +ILS-CSL-soft&0.19{\scriptsize ±0.06}\raisebox{0.5ex}{\scriptsize\textcolor{red}{-41\%}}&0.10{\scriptsize ±0.01}\raisebox{0.5ex}{\scriptsize\textcolor{red}{-33\%}}&0.73{\scriptsize ±0.01}\raisebox{0.5ex}{\scriptsize\textcolor{red}{-1\%}}&0.64{\scriptsize ±0.04}\raisebox{0.5ex}{\scriptsize\textcolor{red}{-12\%}}&0.70{\scriptsize ±0.02}\raisebox{0.5ex}{\scriptsize\textcolor{red}{-15\%}}&0.64{\scriptsize ±0.02}\raisebox{0.5ex}{\scriptsize\textcolor{red}{-17\%}}&0.76{\scriptsize ±0.02}\raisebox{0.5ex}{\scriptsize\textcolor{red}{-4\%}}&0.56{\scriptsize ±0.03}\raisebox{0.5ex}{\scriptsize\textcolor{red}{-3\%}}\\\hline
    \end{tabular}
    \label{tab_cmp_2}
    \end{table*}

\begin{table}[!h]
\centering
\small
\caption{Ranking of methods in Table \ref{tab_cmp_2}.}
\setlength{\tabcolsep}{0.5pt}
\begin{tabular}{cccccc|cccccc}
\hline
\multicolumn{6}{c|}{BDeu}                                                      & \multicolumn{6}{c}{BIC}                                             \\ \hline
\makecell[c]{MIN-\\OBSx} & +hard        & \multicolumn{1}{c|}{+soft}     & HC   & +hard & +soft & \makecell[c]{MIN-\\OBSx} & +hard & \multicolumn{1}{c|}{+soft} & HC   & +hard & +soft \\ \hline
7.0     & \textbf{2.7} & \multicolumn{1}{c|}{{\underline{2.8}}} & 10.1 & 3.4   & 5.3   & 9.8     & 4.9   & \multicolumn{1}{c|}{6.2}   & 11.2 & 6.6   & 8.0   \\ \hline
\end{tabular}
\label{tab_app_rank_2}
\end{table}

Key observations from Table \ref{tab_cmp_ban} are presented as follows.
\begin{enumerate}
    \item ILS-CSL consistently improves the quality of data-based CSL in all cases, with the sole exception observed in the \textit{Cancer} dataset with 250 samples, where it maintains the same performance.
    In contrast, sepLLM shows consistent improvement only in the \textit{Cancer} and \textit{Child} datasets, while exhibiting partial performance degradation in others. This observation underscores the robust and stable enhancement offered by our ILS-CSL framework.
    \item Our framework outperforms sepLLM in datasets with more than 20 variables, albeit showing lesser performance in small-scale datasets, \textit{Cancer} and \textit{Asia}.
This trend is attributed to the relatively simple causal mechanisms in these smaller datasets, where LLM effectively infers correct causal relationships between variables (refer to Table \ref{app_tab_esti} in Section \ref{app_para_estimate}). Despite sepLLM leveraging all existing causality inferred by LLM, its advantage is pronounced only in these two datasets. As the complexity of causal mechanisms increases with the number of variables, the quality of LLM inference diminishes, highlighting the resilience of our  framework against imperfect LLM inference.
\end{enumerate}

Table \ref{tab_app_rank_1} demonstrates that ILS-CSL consistently ranks within the top two positions. Notably, within the sepLLM framework, CaMML, which uses soft constraints, outperforms MINOBSx, which relies on hard constraints. However, this trend reverses in the ILS-CSL framework. This shift is attributed to the ability of soft constraints to filter out some incorrect prior structures that significantly conflict with the data distribution. The prior constraints in sepLLM are not as high-quality as those in LLM-CSL, and the use of ancestral constraints in sepLLM tends to introduce erroneous edges.

\subsection{ILS-CSL With Diverse Backbone Algorithms (RQ2)}

We experiment with varying scoring functions, BDeu and BIC scores, and search algorithms, MINOBSx and HC, and compare to corresponding data-based CSL performances.
Moreover, we experiment with both hard and soft approaches to apply prior constraints, with the prior probability setting $P_{\lambda}=0.99999$ introduced in Equation (\ref{eq_soft_local}).
The results on the utilized observed data of eight datasets are reported in Table \ref{tab_cmp_2}.
The Friedman ranking of the methods is reported in Table \ref{tab_app_rank_2}.
Key observations include:
\begin{table*}[!h]
\small
\setlength{\tabcolsep}{5pt}
\caption{The precision along with ratio of different structures of different answers by GPT-4.}
\begin{center}
\begin{tabular}{c|c||cc|c||cc||c||cc}\hline
 & &   &     &    &     &&    & \multicolumn{2}{c}{Overall Precision}\\ \cline{9-10} 
\multirow{-2}{*}{Answer} & \multirow{-2}{*}{Dataset} & \multirow{-2}{*}{\begin{tabular}[c]{@{}c@{}}Direct \\ edges\end{tabular}} & \multirow{-2}{*}{\begin{tabular}[c]{@{}c@{}}Reversed \\ edges\end{tabular}} & \multirow{-2}{*}{Precision}  & \multirow{-2}{*}{\begin{tabular}[c]{@{}c@{}}Indirect \\ paths\end{tabular}} & \multirow{-2}{*}{\begin{tabular}[c]{@{}c@{}}Reversed \\ indirect paths\end{tabular}} & \multirow{-2}{*}{\begin{tabular}[c]{@{}c@{}}Not \\ reachable\end{tabular}} & \multicolumn{1}{c|}{Qualitative}& Structural \\ \hline \hline
 & Alarm   & 0.33  & 0.02& \cellcolor[HTML]{FADADE}0.94 & 0.28& 0.00 & 0.37   & \multicolumn{1}{c|}{\cellcolor[HTML]{D9E1F4}0.61} & 0.33 \\
 & Asia    & 0.44  & 0.00& \cellcolor[HTML]{FADADE}1.00 & 0.50& 0.00 & 0.06   & \multicolumn{1}{c|}{\cellcolor[HTML]{D9E1F4}0.94} & 0.44 \\
 & Barley  & 0.22  & 0.12& \cellcolor[HTML]{FADADE}0.65 & 0.23& 0.12 & 0.31   & \multicolumn{1}{c|}{\cellcolor[HTML]{D9E1F4}0.45} & 0.22 \\
 & Cancer  & 0.36  & 0.09& \cellcolor[HTML]{FADADE}0.80 & 0.36& 0.09 & 0.09   & \multicolumn{1}{c|}{\cellcolor[HTML]{D9E1F4}0.73} & 0.36 \\
 & Child   & 0.46  & 0.02& \cellcolor[HTML]{FADADE}0.96 & 0.26& 0.04 & 0.22   & \multicolumn{1}{c|}{\cellcolor[HTML]{D9E1F4}0.72} & 0.46 \\
 & Insurance   & 0.41  & 0.05& \cellcolor[HTML]{FADADE}0.89 & 0.32& 0.06 & 0.15   & \multicolumn{1}{c|}{\cellcolor[HTML]{D9E1F4}0.74} & 0.41 \\
 & Mildew  & 0.45  & 0.04& \cellcolor[HTML]{FADADE}0.92 & 0.36& 0.03 & 0.11   & \multicolumn{1}{c|}{\cellcolor[HTML]{D9E1F4}0.82} & 0.45 \\
\multirow{-8}{*}{A}& Water   & 0.47  & 0.13& \cellcolor[HTML]{FADADE}0.78 & 0.11& 0.01 & 0.28   & \multicolumn{1}{c|}{\cellcolor[HTML]{D9E1F4}0.58} & 0.47 \\ \hline \hline
 & Alarm   & 0.02  & 0.36& \cellcolor[HTML]{FADADE}0.95 & 0.10& 0.18 & 0.34   & \multicolumn{1}{c|}{\cellcolor[HTML]{D9E1F4}0.54} & 0.36 \\
 & Asia    & 0.00  & 0.50& \cellcolor[HTML]{FADADE}1.00 & 0.00& 0.36 & 0.14   & \multicolumn{1}{c|}{\cellcolor[HTML]{D9E1F4}0.86} & 0.50 \\
 & Barley  & 0.02  & 0.21& \cellcolor[HTML]{FADADE}0.91 & 0.08& 0.43 & 0.25   & \multicolumn{1}{c|}{\cellcolor[HTML]{D9E1F4}0.64} & 0.21 \\
 & Cancer  & 0.00  & 0.60& \cellcolor[HTML]{FADADE}1.00 & 0.00& 0.00 & 0.40   & \multicolumn{1}{c|}{\cellcolor[HTML]{D9E1F4}0.60} & 0.60 \\
 & Child   & 0.00  & 0.45& \cellcolor[HTML]{FADADE}1.00 & 0.24& 0.12 & 0.18   & \multicolumn{1}{c|}{\cellcolor[HTML]{D9E1F4}0.58} & 0.45 \\
 & Insurance   & 0.02  & 0.59& \cellcolor[HTML]{FADADE}0.97 & 0.02& 0.10 & 0.27   & \multicolumn{1}{c|}{\cellcolor[HTML]{D9E1F4}0.68} & 0.59 \\
 & Mildew  & 0.01  & 0.49& \cellcolor[HTML]{FADADE}0.98 & 0.00& 0.14 & 0.35   & \multicolumn{1}{c|}{\cellcolor[HTML]{D9E1F4}0.64} & 0.49 \\
\multirow{-8}{*}{B}& Water   & 0.03  & 0.51& \cellcolor[HTML]{FADADE}0.94 & 0.29& 0.03 & 0.14   & \multicolumn{1}{c|}{\cellcolor[HTML]{D9E1F4}0.54} & 0.51 \\ \hline \hline
 & Alarm   & 0.00  & 0.00& -  & 0.00& 0.03 & 0.97   & \multicolumn{1}{c|}{0.97} & \cellcolor[HTML]{FADADE}1.00 \\
 & Asia    & 0.00  & 0.00& -  & 0.00& 0.00 & 1.00   & \multicolumn{1}{c|}{1.00} & \cellcolor[HTML]{FADADE}1.00 \\
 & Barley  & - & -   & -  & -   & -    & -   & \multicolumn{1}{c|}{-} & \cellcolor[HTML]{FADADE}- \\
 & Cancer  & - & -   & -  & -   & -    & -  & \multicolumn{1}{c|}{-}  & \cellcolor[HTML]{FADADE}-    \\
 & Child   & 0.00  & 0.11& -  & 0.00& 0.11 & 0.79   & \multicolumn{1}{c|}{0.79} & \cellcolor[HTML]{FADADE}0.89 \\
 & Insurance   & 0.03  & 0.05& -  & 0.00& 0.10 & 0.83   & \multicolumn{1}{c|}{0.83} & \cellcolor[HTML]{FADADE}0.93 \\
 & Mildew  & 0.00  & 0.01& -  & 0.32& 0.36 & 0.32   & \multicolumn{1}{c|}{0.32} & \cellcolor[HTML]{FADADE}0.99 \\
\multirow{-8}{*}{C}& Water   & 0.00  & 0.04& -  & 0.30& 0.19 & 0.47   & \multicolumn{1}{c|}{0.47} & \cellcolor[HTML]{FADADE}0.96\\\hline
\end{tabular}
\end{center}
\label{tab_app_prec}
\end{table*}
\begin{enumerate}
    \item Nearly all scenarios showcase an enhancement, underscoring the impactful role of ILS-CSL in improving CSL performance across diverse datasets and algorithms.
    \item ILS-CSL's impact on causal discovery significantly surpasses the limitations imposed by scoring functions and search algorithms. The ranking results demonstrate this clearly, as HC+ILS-CSL exceeds the performance of MINOBSx, even with a less robust baseline. This also holds true across different scoring functions, highlighting ILS-CSL's broad applicability and effectiveness in improving causal discovery outcomes.
\item The hard approach outperforms the soft approach, attributed to the high quality of specified constraints within ILS-CSL. 
This stands in stark contrast to the findings by \cite{ban2023query}, where the soft approach fared better due to the lower quality of prior constraints.
\end{enumerate}

\begin{figure}
    \centering
    \includegraphics[width=0.48\textwidth]{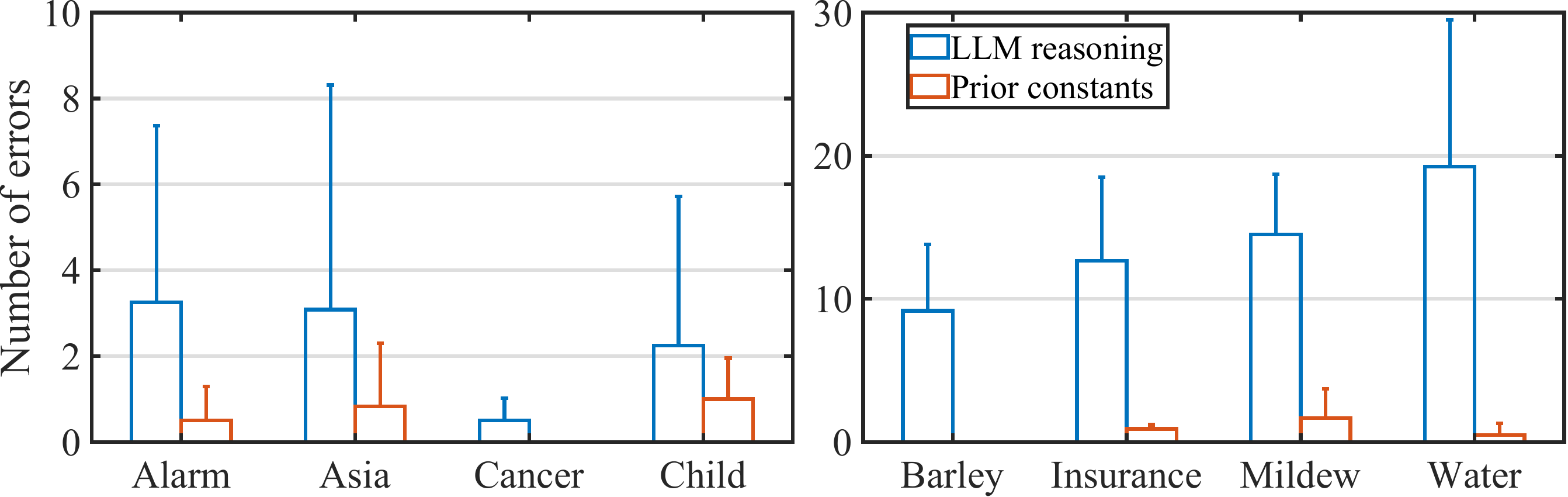}
  \caption{Erroneous LLM inference and erroneous specified edge constraints of MINOBSx-BDeu+ILS-CSL-hard.} 
  \label{fig_errors_min_bdeu}
\end{figure}

\subsection{Errors in LLM Inference and Prior Constraints (RQ3)}

This section is dedicated to the evaluation of ILS-CSL's robustness against the inaccuracies in LLM inference. We scrutinize the erroneous causal relationships inferred by LLM on the edges of the learned DAG, along with the incorrect prior constraints that stem from them. The results pertaining to each dataset, which includes two unique sizes of observed data related to MINOBSx-BDeu with the hard approach, are illustrated in Figure \ref{fig_errors_min_bdeu}. 
For a more comprehensive set of results, refer to the external repository.

Our observations highlight a substantial reduction in the errors of specified edge constraints compared to erroneous LLM inference. This reduction stems from the strategy of only imposing constraints on causality that is inconsistent with what has been learned.
A more detailed analysis on the superior aspect of ILS-CSL to reduce erroneous constraints is made in the following experiment.

\subsection{Why Resistant to Imperfect LLM Inference (RQ3)}
\label{app_sec_est_prec}

This section elucidates the ability of ILS-CSL to minimize prior errors by limiting LLM supervision to edges. We present the ratio of various real structures corresponding to all pairwise variables inferred by GPT-4. Table \ref{tab_app_prec} displays the results for all datasets, highlighting the precision related to ILS-CSL (light red cells) and full inference (light blue cells). It distinguishes between qualitative precision (correct paths) and structural precision (correct edges only).

In the context of the analysis, the outcomes A, B, and C from GPT-4 have specific meanings related to inferred causal relationships between two variables \(X_1\) and \(X_2\):\\
\textbf{Outcome A:} GPT-4 infers that \(X_1\) causes \(X_2\) (\(X_1 \rightarrow X_2\)).\\
 \textbf{Outcome B:} GPT-4 infers that \(X_2\) causes \(X_1\) (\(X_2 \rightarrow X_1\)).\\
\textbf{Outcome C:} GPT-4 infers that \(X_1\) and \(X_2\) are not causally related (\(X_1 \nleftrightarrow X_2\)).\\
In the table, various columns represent the ratio of different corresponding structures in the ground truth:\\
\textbf{Direct Edges:} The edge (\(X_1 \rightarrow X_2\)) exists in truth.\\
\textbf{Reversed Edges:} An reversed edge  (\(X_2 \rightarrow X_1\)) exists in truth.\\
\textbf{Indirect Paths:} A path (\(X_1 \leadsto X_2\)) exists, but (\(X_1 \nrightarrow X_2\)).\\
\textbf{Reversed Indirect Paths:}  (\(X_2 \leadsto X_1\)), but (\(X_2 \nrightarrow X_1\)).\\
\textbf{Not Reachable:} (\(X_1 \not \leadsto X_2, X_2 \not \leadsto X_1\)).

The precision of LLM on variables that have edges (light red cells of answers A and B) is notably high, significantly exceeding the precision on variables that may not. Analyzing prior errors in ILS-CSL reveals:

\begin{enumerate}
    \item For GPT-4 outcome C, the corresponding edge forbidden constraints exhibit high precision, generating few erroneous structural constraints. This is attributed to the high confidence in the absence of causal relations inferred based on knowledge, leading to excellent precision on pairwise variables without structural edges, albeit with a lower recall.
    \item For GPT-4 outcomes A or B, high precision is observed on learned edges belonging to the true skeleton, producing few erroneous structural constraints. Given known direct causality between pairwise variables, LLM can easily infer the correct causal direction, stemming from the counterintuitive nature of reversed causal statements.
    \item Major LLM inference errors stem from outcomes A and B on learned edges outside the true skeleton. However, the impact of these errors on generating incorrect structural constraints is mitigated by the low probability of extra edges occurring in a learned structure ($z_3\approx 0.07$, see Table \ref{app_tab_strutural_para}) and the strategy of specifying a prior constraint only when inconsistent.
    \label{point_3}
\end{enumerate}

In essence, the primary limitation of LLM in causal inference is the confusion between direct causal relationships, indirect causality, and correlations, evidenced by the low overall qualitative and structural precision. This limitation hampers the performance of using LLM-derived existence on causality as ancestral (qualitative precision) or edge constraints (structural precision) seperately.

Contrarily, ILS-CSL effectively minimizes prior errors by leveraging the inherent precision of LLM in inferring non-causal relations and determining causal direction on pairwise variables with direct causality. It smartly circumvents LLM's limitation in discerning the existence of direct causal relationships, which are easily confused with indirect causality or correlations, by restricting the LLM inference into the range of learned structures from data, as analyzed in point \ref{point_3}.


\begin{figure}[!h]
 \centering
\includegraphics[width=.48\textwidth]{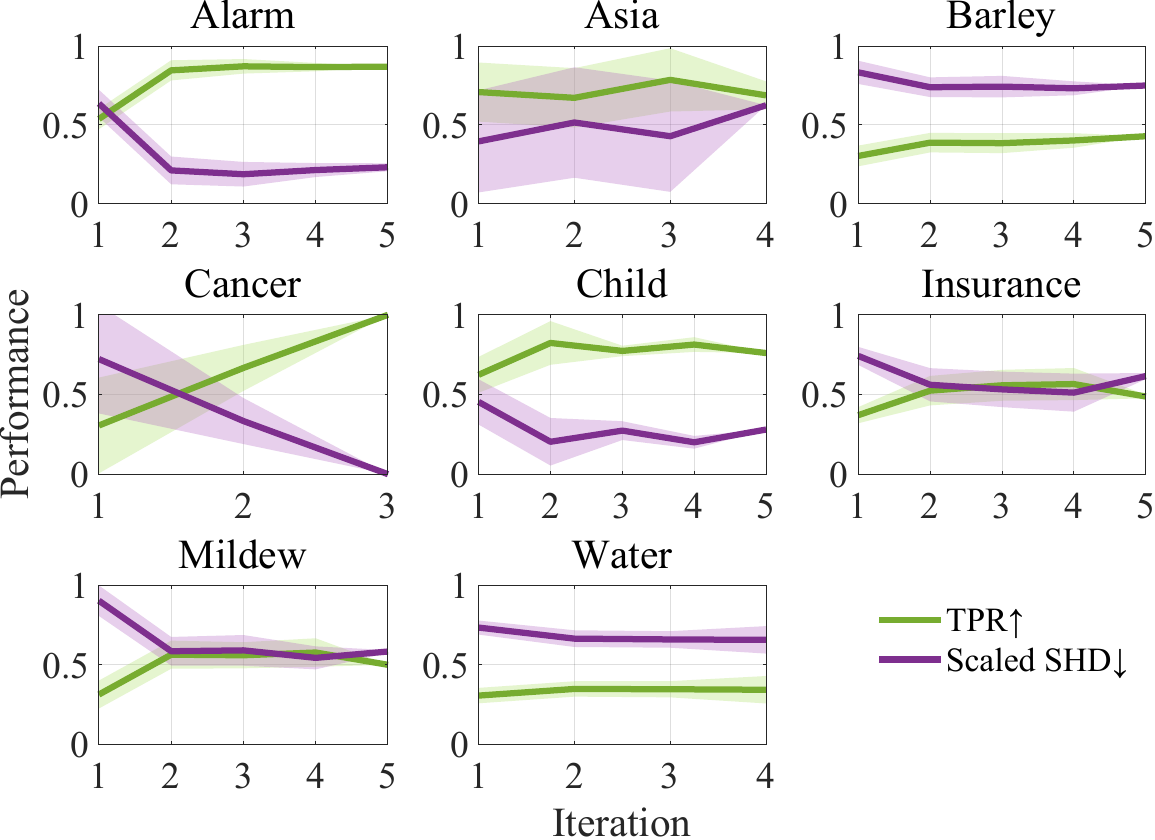}
 \caption{Trend of TPR$\uparrow$ (green line) and scaled SHD$\downarrow$ (purple line) of HC+BIC+ILS-CSL-hard on various datasets.}
 \label{fig_app_iter_shd}
\end{figure}

\subsection{Trend of DAG Quality over Iterations (RQ4)}

This section outlines the iterative trends of scaled SHD (aiming for a decrease, denoted as SHD$\downarrow$) and True Positive Rate (aiming for an increase, denoted as TPR$\uparrow$) for various backbone algorithms across eight datasets. Each dataset spans two distinct data sizes, resulting in 12 segments of observed data. It's crucial to note the potential for significant derivation due to performance differences across varying data sizes, particularly for smaller-scale datasets like \textit{Cancer} and \textit{Asia}.
The results of HC+BIC+ILS-CSL-hard on various datasets are reported in Figure \ref{fig_app_iter_shd}, with comprehensive results available in the external repository.
Key observations from the iterative trends include:
\begin{itemize}
    \item \textbf{Limited Iteration Numbers:} Most cases require a limited number of iterations. The area near the maximum iteration in each figure is small when exceeding 4, indicating that few out of the 12 cases reach this point. Some cases even have a derivation of zero at the maximum iteration, signifying that only one case attains this maximum value.
    \item \textbf{Quality Improvement Trend:} Generally, as the iteration number increases, the scaled SHD decreases, and the TPR increases. This trend underscores the enhancement in the quality of the learned causal structures as ILS-CSL progresses.
    \item \textbf{Significant Initial Improvement:} The most substantial improvement in the quality of learned causal DAGs occurs in the first round of LLM supervision (from Iteration 1 to 2). Subsequent iterations offer diminished enhancements. This pattern is attributed to the initial presentation of most inconsistent edges with LLM inference in the first iteration. Post the integration of prior constraints, the new structures learned by CSL exhibit far fewer inconsistencies with LLM inference.
    \item \textbf{Potential Quality Degradation:} In certain instances, the quality of the causal DAG diminishes across specific iterations. This decline could stem from the introduction of new erroneous prior constraints in a given iteration or a statistical artifact. The latter scenario arises when two consecutive iterations do not employ the same set of observed data, as some cases conclude in the preceding iteration.
\end{itemize}

These observations provide a comprehensive insight into the iterative behavior of ILS-CSL, highlighting its effectiveness and areas of caution to ensure consistent enhancement in learned causal structures.

\begin{figure*}[!h]
 \centering
\includegraphics[width=1.0\textwidth]{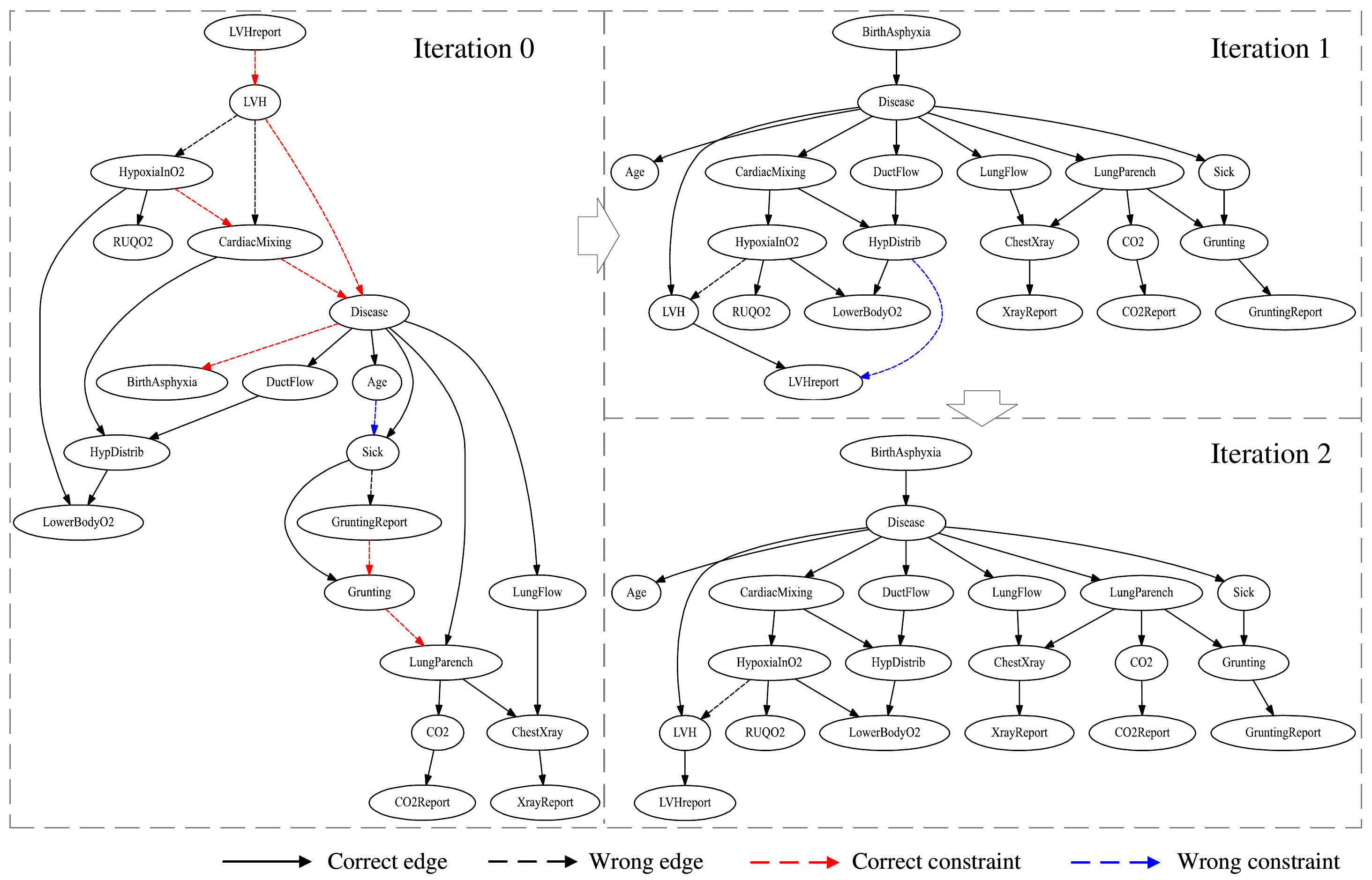}
 \caption{Visualized process of HC-BDeu+ILS-CSL-hard on a set of observed data of \textit{Child}, 2000 samples.
 The SHD of iterations are: 12 for Iteration 0, 3 for Iterations 1 and 2.}
 \label{fig_app_visual}
\end{figure*}

\subsection{Illustrative Example of DAG Evolution (RQ4)}
We visualize the learned causal structures in iterations to unfold the details of ILS-CSL.
An illustrative example by HC (BDeu) algorithm on \textit{Child} dataset, 2000 samples, with hard constraining approach in ILS-CSL, is reported in Figure \ref{fig_app_visual}.

Initially, HC (BDue) learns a causal DAG from pure observed data (Iteration 0), whose edges are supervised by LLM, leading to edge constraints (colored arrows) on inconsistent inferred edge by LLM.
The constraints could refine local structures (red arrows) or bring harm due to the erroneous inference (blue arrows).
The erroneous edges (dotted arrows) are reduced as the iteration goes. 
Details of further observations are presented as follows:
\begin{itemize}
 \item The SHD of the learned causal DAG is greatly reduced from 12 to 3 by employing the ILS-CSL framework, showcasing the significant capability of our framework to enhance the quality of learned causality.
 \item The first round of LLM-based supervision refines the learned DAG to a much greater extent than the following rounds.
 This addresses the acceptable efficiency loss of ILS-CSL, which usually does not require many iterations.
 \item There are $7$ correct constraints (red arrow) and $2$ erroneous ones (blue arrow) in total.
 The number of directly corrected edges by these priors is $7-2 = 5$, while the reduced SHD is $8$, meaning that 3 edges that are distinct from those in constraints are corrected without any prior knowledge on them.
 It underscores the capability of discovering structures unrelated to prior constraints by integrating them.
 This phenomenon could be interpreted as the capability of aiding discovery of unknown causal mechanisms by the known causal knowledge.
\end{itemize}


\section{Conclusions}
This paper presents ILS-CSL, a framework that enhances causal discovery from data using Large Language Models (LLMs). ILS-CSL seamlessly incorporates LLM inference on the edges of the learned causal Directed Acyclic Graph (DAG), converting qualitative causal statements into precise edge-level prior constraints while effectively mitigating constraint errors stemming from imperfect prior knowledge. Comprehensive experiments across eight real-world datasets demonstrate the substantial and consistent improvement ILS-CSL brings to the quality of causal structure learning (CSL) outputs. 
Notably, ILS-CSL surpasses the existing separate way to guide CSL by applying LLM inferred causality as ancestral constraints, with a marked performance increase as the number of variables grows. 
This advancement underscores the promising application of the ILS-CSL framework in assistance of complex, real-world causal discovery tasks.



\newpage

\bibliography{ref}
\bibliographystyle{IEEEtran}

\end{document}